\documentclass[runningheads]{llncs}

\usepackage[final]{eccv}

\usepackage{eccvabbrv}

\usepackage{booktabs}
\usepackage{multirow}
\usepackage{graphicx}
\usepackage[table]{xcolor}
\usepackage[normalem]{ulem}
\usepackage{caption}
\usepackage{wrapfig}

\newcommand{\best}[1]{\textbf{\textcolor{red}{#1}}}
\newcommand{\second}[1]{\textbf{\textcolor{blue}{#1}}}

\definecolor{cvprred}{RGB}{250, 235, 235}
\definecolor{cvprgray}{RGB}{240, 240, 240}
\definecolor{loraorange}{RGB}{255, 165, 0}
\definecolor{txtpurple}{RGB}{240, 230, 250} 
\useunder{\uline}{\ul}{}

\usepackage[accsupp]{axessibility}  

\usepackage{hyperref}
\hypersetup{
    colorlinks=true,
    linkcolor=eccvblue,
    citecolor=eccvblue,
    urlcolor=eccvblue
}

\usepackage{orcidlink}

\begin{document}

\title{Pro-Pose: Unpaired Full-Body Portrait Synthesis via Canonical UV Maps} 

\titlerunning{Pro-Pose}

\author{
Sandeep Mishra$^{1*\dagger}$ \quad
Yasamin Jafarian$^2$ \quad
Andreas Lugmayr$^2$ \\
Yingwei Li$^2\ddagger$ \quad
Varsha Ramakrishnan$^2$ \quad
Srivatsan Varadharajan$^2$ \\
Alan C. Bovik$^{1}$ \quad
Ira Kemelmacher-Shlizerman$^{2}$
\vspace{0.2cm} \\ 
$^1$The University of Texas at Austin, USA \quad $^2$Google, USA \\
{\tt\small \url{https://pro-pose-portrait.github.io}} \\
{\tt\small sandy.mishra@utexas.edu, bovik@ece.utexas.edu} \\
{\tt\small \{jafarian, alugmayr, yingweili, vio, srivatsanv, kemelmi\}@google.com}
}

\authorrunning{S.~Mishra et al.}

\institute{}

\onecolumn{%
\renewcommand\onecolumn[1][]{#1}%
\maketitle
\begin{center}
    \captionsetup{type=figure}
    \vspace{-4mm}
    \includegraphics[width=\linewidth]{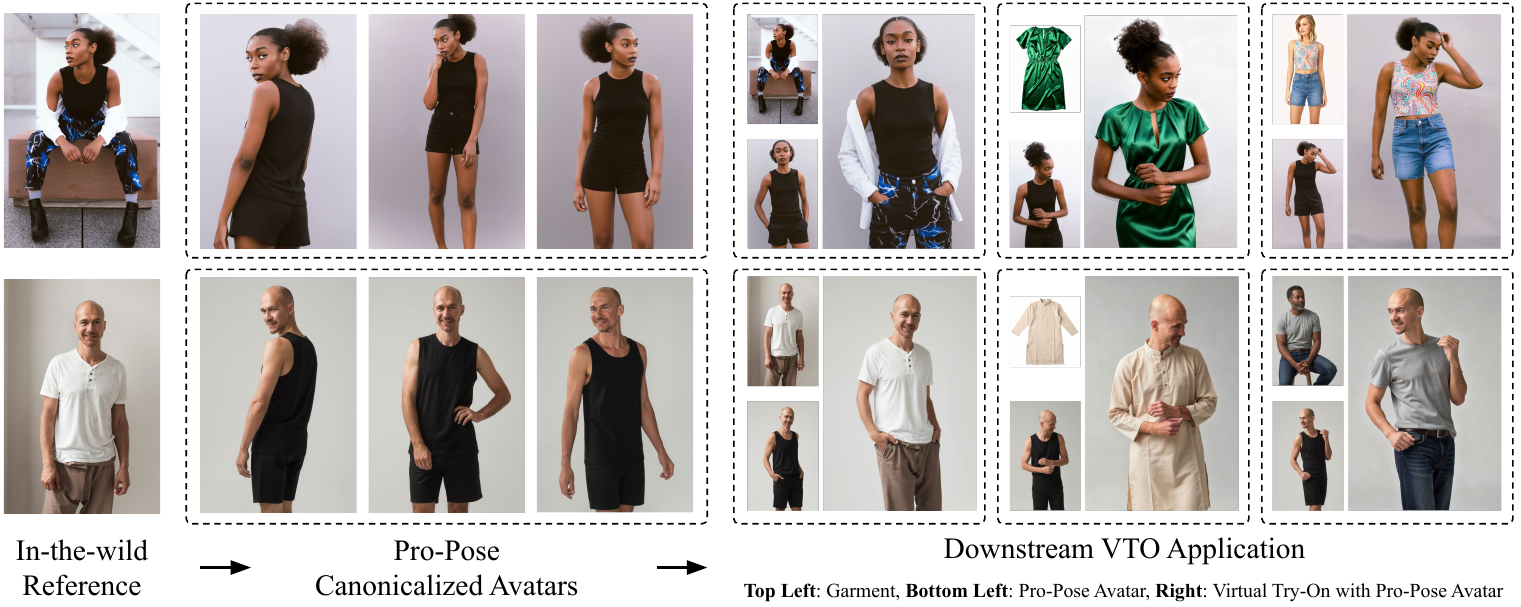}
    \vspace{-5mm}
    \captionof{figure}{\textbf{Pro-Pose Canonicalization and Downstream Virtual Try-On.} Given a single in-the-wild photo (left), Pro-Pose synthesizes high-fidelity, pose-controllable avatars (middle) driven by arbitrary SMPL-X~\cite{Pavlakos2019SMPLX} poses. By canonicalizing the subject into a minimal black outfit, our method preserves the user's visible identity -- facial features, skin textures, and body shape, while stripping away the occlusions of the original clothing. This standardized representation provides an optimal, clean geometric canvas for downstream tasks like Virtual Try-On (right), enabling off-the-shelf VTO~\cite{googlevto2023} models to apply novel garments without interference from the source image.}
    \label{fig:teaser}
    \vspace{-4mm}
\end{center}%
}

\begin{NoHyper}
{
  \let\thefootnote\relax\footnotetext{%
  \footnotesize
  \textsuperscript{$*$}Work done during an internship at Google.
  \textsuperscript{$\dagger$}Corresponding author.
  \textsuperscript{$\ddagger$}Yingwei Li was affiliated with Google at the time of this research.
  }
}
\end{NoHyper}

\begin{abstract}

Photographs of people taken by professional photographers typically present the person in beautiful lighting, with an interesting pose, and flattering quality. This is unlike common photos people take of themselves in uncontrolled conditions. In this paper, we explore how to canonicalize a person's "in-the-wild" photograph into a controllable, high-fidelity avatar---reposed in a simple environment with standardized minimal clothing. A key challenge is preserving the person’s unique whole-body identity, facial features, and body shape while stripping away the complex occlusions of their original garments. While a large paired dataset of the same person in varied clothing and poses would simplify this, such data does not exist. To that end, we propose two key insights: 1) Our method transforms the input photo into a canonical full-body UV space, which we couple with a novel reposing methodology to model occlusions and synthesize novel views. Operating in UV space allows us to decouple pose from appearance and leverage massive unpaired datasets. 2) We personalize the output photo via multi-image finetuning to ensure robust identity preservation under extreme pose changes. Our approach yields high-quality, reposed portraits that achieve strong quantitative performance on real-world imagery, providing an ideal, clean biometric canvas that significantly improves the fidelity of downstream applications like Virtual Try-On (VTO).

\end{abstract}    
\vspace{-10mm}

\section{Introduction}
\label{sec:intro}

Images have become the foundation of digital identity, yet everyday photos are often captured under uncontrolled conditions with cluttered backgrounds, imperfect lighting, and arbitrary poses and clothing. Consequently, obtaining a clean, controllable ``photoshoot-style'' representation is difficult, primarily due to scarce high-quality training data. Existing reposing methods such as MCLD~\cite{MCLD} and LEFFA~\cite{Zhou2025LEFFA} rely on the small-scale paired DeepFashion dataset~\cite{deepfashion} ($\approx$100 identities), leading to overfitting and limited generalization to unseen individuals, and they cannot modify garments. In contrast, virtual try-on (VTO) models~\cite{Han2018VITON} typically rely on paired data of a source person and a target garment. Crucially, during training, they derive the person input from the target image and thus are trained to preserve the original pose rather than control or modify it. Consequently, neither method can jointly repose a person and generalize to real-world imagery - motivating our goal to generate photorealistic, reposed portraits that preserve {full-body identity (face, body shape, and skin features) in standardized conditions. {Beyond generating standalone digital avatars, this controlled canonical representation acts as a powerful intermediate foundation for downstream tasks like VTO. Current VTO models often suffer from unwanted structural leakage or geometric guidance from the subject's original clothing, which frequently leads to failures on unconstrained in-the-wild photos. By deliberately canonicalizing the subject into a minimal black tank top and shorts, Pro-Pose provides VTO pipelines with an effective, form-fitting 'blank canvas' that robustly preserves body shape and pose while eliminating the geometric interference of source garments. As shown in Figure~\ref{fig:teaser}, our approach produces a versatile portfolio of reposed, identity-faithful portraits from a single input.

In this paper, we address the limited identity diversity by leveraging both abundant single images~\cite{ffhq, Zhu2024MMVTO, Cheng2023TryOnDiffusion} and scarce paired data~\cite{deepfashion}. We propose a self-supervised framework utilizing a canonical UV-space~\cite{Pavlakos2019SMPLX}, which theoretically decouples pose from texture. However, practical extraction remains ill-posed, since the boundaries of the visible texture map inevitably leak pose information. To eliminate these clues, we introduce \textit{Donor-based UV Reposing}, which uses occlusion masks from unrelated ``donor'' images to disguise the original visibility patterns, forcing the model to learn robust geometric warping. This formulation enables seamless joint training on 30K paired samples and $\approx$470K unpaired single images. Finally, we employ Gemini 2.5 Flash Image~\cite{nanobanana} to standardize the garments in the target images, creating a \textbf{B}ase \textbf{C}lothing (\textbf{BC}) dataset featuring subjects in black tops and shorts {without altering their identity or pose.

Our method employs a dual-branch training strategy to leverage both paired and unpaired data. The paired branch conditions the model on a SMPL-X~\cite{Pavlakos2019SMPLX} target, the partial UV texture map, and a source face crop to synthesize the subject in the standardized base garment. Conversely, for unpaired data, we utilize a self-supervised approach where the input also serves as the target. Here, we extract a donor-based UV map and drop out the face condition. This forces the network to reconstruct identity solely from the warped texture, preventing it from trivially copying input pixels. Finally, to address the overfitting risks inherent in single-shot personalization methods trained on limited data, we introduce a lightweight finetuning strategy. This adapts the pre-trained priors to new subjects using minimal reference images, ensuring high identity preservation without requiring large-scale multi-image datasets.

We summarize our contributions as follows: (1) a scalable joint training framework that integrates scarce paired data~\cite{deepfashion} with abundant unpaired single images~\cite{ffhq, Zhu2024MMVTO, Cheng2023TryOnDiffusion} to overcome the limited identity diversity of existing benchmarks; (2) a novel self-supervised Donor-based UV Reposing mechanism that reduces coupling between pose and {full-body texture to prevent boundary-based pose leakage, enabling effective learning from unpaired data; (3) a {strategy to create a large-scale \textbf{B}ase \textbf{C}lothing (\textbf{BC}) dataset~\cite{deepfashion, ffhq}, synthesized via Gemini 2.5 Flash Image~\cite{nanobanana} to ensure consistent visual conditions; (4) a subject-specific adaptation mechanism that utilizes our unified training objective to fine-tune {LoRA layers at test-time, significantly reducing identity drift in challenging scenarios; and {(5) extensive evaluations - including comparisons against Gemini 2.5 Flash Image~\cite{nanobanana}, Gemini 3 Pro Image~\cite{nanobananapro} and demonstrations of downstream VTO - showing that our approach achieves state-of-the-art, identity-faithful synthesis on challenging real-world imagery.

\vspace{-3mm}
\section{Related Work}
\label{sec:related_work}
\vspace{-2mm}

\noindent \textbf{2D Human Image Synthesis.} Research in 2D human image synthesis has historically split into related tracks: pose-guided generation and disentangled attribute editing. Pose-guided methods aim to generate a novel pose of a person from a single image~\cite{Siarohin2018DefGAN,Zhu2019PATN,Li2019DIAF,Ren2020GFLA}. To this aim, PG$^2$~\cite{Ma2017PG2} adopted a coarse-to-fine pipeline, while subsequent single-stage architectures improved end-to-end training by incorporating spatial alignment via deformable skip connections~\cite{Siarohin2018DefGAN}, progressive attention~\cite{Zhu2019PATN}, or dense appearance flow~\cite{Li2019DIAF}. Parallel work focused on disentangling attributes, such as appearance and shape~\cite{Esser2018VAE}, or attribute-specific editing~\cite{Men2020ADGAN,Park2021PISE}. This line of work was further generalized by motion models for arbitrary object animation~\cite{Siarohin_2019_PAMI}. To reduce paired data needs, \cite{Sanyal2021SPICE} introduced cyclic self-supervision. However, these 2D models struggle with spatial consistency and entangle pose with texture, leading to artifacts and identity loss.

\noindent \textbf{3D-Aware Human Generation.} To overcome spatial inconsistency, later methods incorporated 3D reasoning, implicit surface modeling, or neural rendering~\cite{Saito2019PIFu,Saito2020PIFuHD,Huang2020ARCH,Xiu2022ICON,Weng2022HumanNeRF}. 3DHumanGAN~\cite{Yang2023ThreeDHumanGAN} introduced a 3D-pose mapping module to enforce geometric plausibility, and EG3D~\cite{Chan2022EG3D} established efficient hybrid explicit/implicit 3D GANs for view-consistent generation. Other work such as PIFu~\cite{Saito2019PIFu} and PIFuHD~\cite{Saito2020PIFuHD} recovered pixel-aligned implicit fields, while ConTex-Human~\cite{Gao2024ConTexHuman} focused on achieving texture consistency. More recent approaches have explored 3D Gaussian Splatting for real-time avatars~\cite{kerbl3Dgaussians,moreau2023human,moreau2024human} and self-supervised learning from social media videos~\cite{Jafarian_2022_TPAMI,Jafarian_2021_CVPR_TikTok}. However, these approaches rely on strict multi-view or 3D supervision. While such datasets may be voluminous in terms of total frames, they remain severely limited in identity diversity compared to 2D images, preventing generalization to the distribution of real-world human appearances.

\vspace{1mm}
\noindent \textbf{Diffusion-Based Animation.} Recent diffusion-based techniques have significantly advanced fidelity and controllability~\cite{CFLD,MCLD,Zhou2025LEFFA,Hu2024AnimateAnyone,Bar2023DisCo,Lugmayr_2022_CVPR,brooks2022instructpix2pix}. Coarse-to-Fine Latent Diffusion (CFLD)~\cite{CFLD} separates semantic and texture modeling; Animate Anyone~\cite{Hu2024AnimateAnyone} and DisCo~\cite{Bar2023DisCo} extend this paradigm to temporal consistency and disentangled control; and follow-up works such as MCLD~\cite{MCLD} and LEFFA~\cite{Zhou2025LEFFA} enhance pose controllability. While high-quality, these models remain highly data-hungry, relying on large paired datasets like DeepFashion. Consequently, they are prone to overfitting and ``identity drift'' on unseen individuals.

\vspace{1mm}
\noindent \textbf{Virtual Try-On and Garment Manipulation.} Virtual try-on (VTO) methods transfer garments onto target bodies while preserving identity~\cite{Han2018VITON,Wang2018CPVTON,Yang2020ACGPN,Lee2022HRVITON,Morelli2022DressCode,Wang2023OOTDiffusion,Zhu2024MMVTO,Cheng2023TryOnDiffusion}. VITON~\cite{Han2018VITON} and CP-VTON~\cite{Wang2018CPVTON} pioneered geometric matching for in-shop clothing. M\&M VTO~\cite{Zhu2024MMVTO} extends this with multi-garment diffusion control. However, these methods assume garment–person pairs rather than person-pose pairs, limiting reposing ability.

\vspace{1mm}
\noindent \textbf{Summary.} Existing 2D models lack consistency, 3D methods require expensive supervision, and diffusion animators rely on scarce paired data. To address these limitations, we propose a self-supervised framework formulated in UV space. By decoupling pose from texture and introducing donor-based reposing, we leverage abundant unpaired data alongside scarce paired data to generate photorealistic, reposed, and garment-neutral avatars from single images.
\vspace{-2mm}
\section{Method}
\label{sec:method}

We address the challenge of synthesizing highly realistic and pose-controllable human avatars by proposing a unified framework capable of learning from both abundant single-view images and scarce paired data. This integration is achieved through a novel supervision strategy formulated directly in canonical UV texture space, which naturally disentangles pose from appearance.

\subsection{Overview and Problem Formulation}
The fundamental goal is to generate an image, $\mathbf{A}_{\mathbf{p}}$, of a specific person in a novel target pose $\mathbf{p}$ (represented as a SMPL-X~\cite{Pavlakos2019SMPLX} rendering),  \textit{standardized into minimal Base Clothing (BC) -- a black sleeveless tank top and shorts}. This canonicalization strips away source-garment occlusions while preserving the subject's identity, body shape, and skin appearance. Given a single reference image $\mathbf{I}_{\mathbf{p}'}$ of that person in a source pose $\mathbf{p}'$, we formulate this as a conditional generation problem:

\begin{equation}
    \label{eq:image-space-obj}
    \mathbf{A}_{\mathbf{p}} = f_\theta(\mathbf{I}_{\mathbf{p}'}, \mathbf{p})
\end{equation}
where $f_\theta$ is the parameterized model we aim to learn.

A standard approach to train such a model involves minimizing a reconstruction loss, such as $\| \mathbf{A}_{\mathbf{p}} - \mathbf{I}_{\mathbf{p}} \|$, where $\mathbf{I}_{\mathbf{p}}$ is a ground-truth image of the same person in the target pose $\mathbf{p}$. This requires a large dataset of pose-paired images $(\mathbf{I}_{\mathbf{p}'}, \mathbf{I}_{\mathbf{p}})$. However, collecting such data at scale presents significant challenges due to scarcity, limited identity diversity, and inconsistencies (e.g. clothing changes) between pairs.

These limitations motivate leveraging abundant unpaired single images. However, naively setting $\mathbf{p} = \mathbf{p}'$ leads to a trivial identity function. The central challenge is therefore to formulate a self-supervised manner that forces the model to learn meaningful representations from a single image.

\subsection{Self-Supervision in Canonical UV Space}
\label{sec:uv_supervision}

To enable effective learning from abundant single-view images, we shift our representation from image space to a canonical UV texture space. Ideally, we would extract a complete, identity-specific texture map $\mathbf{T}$ from any input image, perfectly decoupled from the subject's pose. This would allow us to formulate a unified generator $g_\theta$ for any target pose $\mathbf{p}$:
\begin{equation}
    \label{eq:ideal-obj}
    \mathbf{A}_{\mathbf{p}} = g_\theta(\mathbf{T}, \mathbf{p})
\end{equation}

In practice, single-view unwrapping is ill-posed due to self-occlusions. We can only obtain a \textit{partial} texture map, $\mathbf{T}_{\mathbf{p}} = \mathbf{T} \odot \mathbf{M}_{\mathbf{p}}$, where $\mathbf{M}_{\mathbf{p}} \in \{0, 1\}^{H \times W}$ is the binary visibility mask uniquely defined by pose $\mathbf{p}$ (see Figure~\ref{fig:pose_leakage} a). We must therefore approximate the generator using partial information:
\begin{equation}
    \label{eq:self-sup-obj}
    \mathbf{A}_{\mathbf{p}} \approx g_\theta(\mathbf{T}_{\mathbf{p}}, \mathbf{p})
\end{equation}

\paragraph{Pose Leakage via Occlusion Boundaries.}
While Eq.~\ref{eq:self-sup-obj} enables self-supervision, it introduces a critical challenge: the partial texture $\mathbf{T}_{\mathbf{p}}$ is highly correlated with the pose $\mathbf{p}$ via the mask $\mathbf{M}_{\mathbf{p}}$. As visualized in Figure~\ref{fig:pose_leakage} a, occlusion boundaries in $\mathbf{T}_{\mathbf{p}}$ perfectly align with the target pose. This allows the network to minimize reconstruction loss trivially by "copy-pasting" visible pixels based on these boundaries, rather than learning the desired 3D geometric warping.

\subsubsection{Donor-Based UV Reposing}
\label{sec:UVreposing}

To mitigate pose leakage, we must break the correlation between the input texture boundaries and the target pose. While rendering the textured SMPL-X mesh into novel poses generates pseudo-pairs in image space (Figure~\ref{fig:pose_leakage} b), this process is computationally prohibitive for online training. We instead bypass explicit rendering by formulating the problem directly in UV space via \textit{Donor-based UV Reposing} (Figure~\ref{fig:pose_leakage} c), an efficient 2D strategy that synthetically simulates novel pose occlusions.

\begin{wrapfigure}{r}{0.49\textwidth}
  \centering
  \vspace{-\intextsep}
  \includegraphics[width=\linewidth]{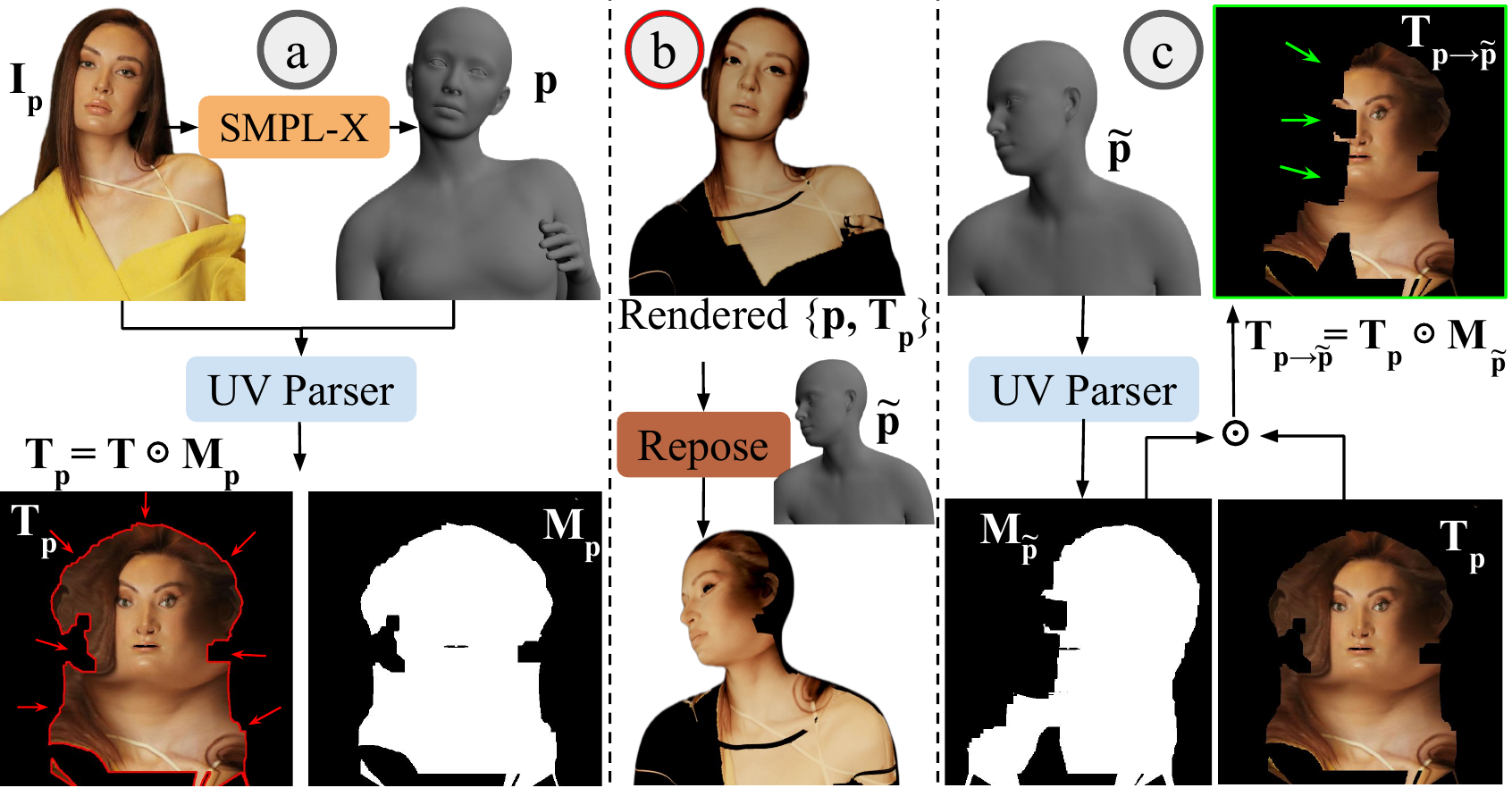}
  \caption{\textbf{Pose Leakage Mitigation.} (a) Standard partial textures $\mathbf{T_p}$ leak source pose information via occlusion boundaries, allowing trivial reconstruction shortcuts. (b) Generating pseudo-pairs via image-space rendering is computationally prohibitive for online training. Moreover, SMPL-X lacks fine details—such as hair—leading to unrealistic renderings. (c) Our \textbf{Donor-based Reposing} efficiently bypasses rendering by applying random donor masks $\mathbf{M}_{\tilde{\mathbf{p}}}$ directly in UV space. This simulates novel occlusions, preventing leakage and forcing the network to learn robust geometric warping. Note: To clearly visualize the subtle boundary differences (indicated by arrows), here we display a crop of just the face region from the full texture map; in practice, donor-reposing applies to all body parts.}
  \label{fig:pose_leakage}
\end{wrapfigure}

We mask the input texture using visibility mask $\mathbf{M}_{\tilde{\mathbf{p}}}$ from a random `donor' image with pose $\tilde{\mathbf{p}}$. Because the UV parameterization is canonical across all subjects, this donor mask can be sourced from \textit{any} other person in the training set, rather than being synthetically generated; in practice we sample donors whose visibility masks share a $40$--$80\%$ overlap (IoU) with the source (Sec.~\ref{sec:implementation}). This yields a hybrid texture $\mathbf{T}_{\mathbf{p} \rightarrow \tilde{\mathbf{p}}} = \mathbf{T}_{\mathbf{p}} \odot \mathbf{M}_{\tilde{\mathbf{p}}}$. Mathematically, this represents the source--donor visibility intersection: $\mathbf{T} \odot (\mathbf{M}_{\mathbf{p}} \odot \mathbf{M}_{\tilde{\mathbf{p}}})$. Because this intersection is commutative, the resulting texture boundaries no longer uniquely characterize the source pose $\mathbf{p}$. This ambiguity prevents trivial boundary-based leakage, forcing the generator to recover the original view via geometric inpainting:
\begin{equation}
    \label{eq:donor-obj}
    \mathbf{A}_{\mathbf{p}} \approx g_\theta(\mathbf{T}_{\mathbf{p} \rightarrow \tilde{\mathbf{p}}}, \mathbf{p})
\end{equation}

\vspace{-5mm}
\subsection{Generative Framework}
\label{sec:framework}

We parameterize our generator $g_\theta$ as a Latent Rectified Flow model~\cite{lipman2023flowmatchinggenerativemodeling}, instantiated using the state-of-the-art Flux.1 [dev] transformer backbone~\cite{flux2024}. Consistent with this architecture, we operate in the compressed 16-channel latent space defined by the pre-trained Flux Variational Autoencoder (VAE), comprising an encoder $\mathcal{E}$ and a decoder $\mathcal{D}$.

Target images $\mathbf{I}_{\mathbf{p}}$ are compressed into latent representations $\mathbf{x}_0 = \mathcal{E}(\mathbf{I}_{\mathbf{p}})$. Following the Flow Matching formulation~\cite{lipman2023flowmatchinggenerativemodeling}, we model the generative process as an Ordinary Differential Equation (ODE)~\cite{NeuralODE} that transports samples from a standard Gaussian noise distribution $\mathbf{x}_1 \sim \mathcal{N}(\mathbf{0}, \mathbf{I})$ to the data distribution $\mathbf{x}_0$.

We train a velocity prediction network $v_{\theta}$ to estimate the \textit{vector field} driving this transport using the standard Flow Matching objective: 
\begin{equation}
\label{eq:fm_loss}
\mathcal{L}_{\text{FM}} = \mathbb{E}_{t, \mathbf{x}_0, \mathbf{x}_1, \mathbf{c}} \left[ \| v_{\theta}(\mathbf{x}_t, t, \mathbf{c}) - (\mathbf{x}_1 - \mathbf{x}_0) \|^2 \right]
\end{equation}
where $t \sim \mathcal{U}[0,1]$ is the timestep, $\mathbf{c}$ is the condition set, and $\mathbf{x}_t = t \cdot \mathbf{x}_1 + (1-t) \cdot \mathbf{x}_0$ represents the linear interpolation between noise ($\mathbf{x}_1$) and data ($\mathbf{x}_0$).

Once trained, our high-level generator $g_\theta(\mathbf{c})$ (Eq.~\ref{eq:donor-obj}) is defined by numerically integrating this field backwards from noise ($t=1$) to data ($t=0$), followed by decoding:
\begin{equation}
    g_\theta(\mathbf{c}) = \mathcal{D} \left( \mathbf{x}_1 + \int_{1}^{0} v_{\theta}(\mathbf{x}_t, t, \mathbf{c}) dt \right)
\end{equation}

\textbf{Conditioning Architecture.} Following OminiControl~\cite{tan2025ominicontrol}, we inject the condition set $\mathbf{c}$ by concatenating the tokenized conditions (UV texture, SMPL-X pose render, and face crop) with the noisy image and text tokens along the sequence dimension, allowing them to interact through the DiT's multi-modal attention. We keep the pre-trained image- and text-stream parameters frozen and apply a LoRA mask so that updates only affect the condition tokens. We additionally inject rank-$128$ LoRA adapters into the attention projections ($\mathbf{Q}, \mathbf{K}, \mathbf{V}, \mathbf{O}$) and the MLP layers of all DiT blocks. This adapts the frozen Flux.1 [dev] backbone to our task while training only a small fraction of the parameters.

\subsection{Training Data and Conditioning Strategy}
\label{sec:data_and_conditioning}

We enable joint training via a unified condition vector $\mathbf{c} = \{ \mathbf{T}_{\text{in}}, \mathbf{p}_{\text{target}}, \mathbf{I}^{FC}_{\text{in}} \}$, comprising the input texture, target pose, and an optional identity-anchoring face crop extracted via MediaPipe~\cite{MediaPipe}. An overview of our unified training framework is presented in Figure~\ref{fig:method_overview}. We instantiate $\mathbf{c}$ based on the data source:

\subsubsection{Paired Supervision}
For datasets with ground-truth pairs (reference $\mathbf{I}_{\mathbf{p}'}$, target $\mathbf{I}_{\mathbf{p}}$), we condition on the reference partial texture and face crop to reconstruct the target view ($\mathbf{x}_0 = \mathcal{E}(\mathbf{I}_{\mathbf{p}})$):
\begin{equation}
    \mathbf{c}_{\text{paired}} = \{ \mathbf{T}_{\mathbf{p}'}, \mathbf{p}, \mathbf{I}^{FC}_{\mathbf{p}'} \}
\end{equation}
The face crop $\mathbf{I}^{FC}_{\mathbf{p}'}$ is included to boost fidelity in facial regions where the partial texture $\mathbf{T}_{\mathbf{p}'}$ may lack resolution.

\begin{figure*}[t]
  \centering
  \vspace{-2mm}
  \includegraphics[width=\linewidth]{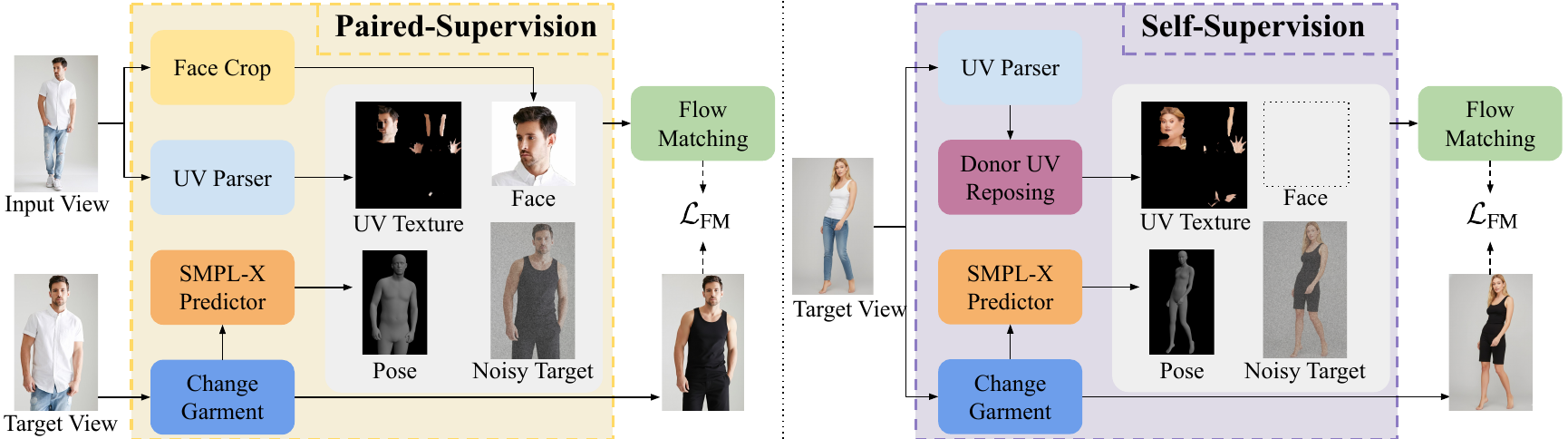}
  \vspace{-3pt}
  \caption{\textbf{Overview of our Avatar Generation Framework.} 
  Our approach leverages single-view datasets by operating in a canonical UV space, extracting UV texture and pose \cite{Pavlakos2019SMPLX}.
  \textbf{Left (Paired Supervision):} When ground-truth pose pairs are available, we condition the Flow Matching model on the partial UV texture, target pose, and face crop.
  \textbf{Right (Single-View Self-Supervision):} To prevent "pose leakage" from occlusion boundaries when training on single images, we introduce a \textbf{Donor-based UV Reposing} module (Sec.~\ref{sec:UVreposing}). This synthetically re-poses the input texture using a random donor visibility mask, forcing the model to learn robust identity representations. Furthermore, we drop-out the face crop condition in this branch to prevent trivial reconstruction via pixel-perfect information leakage.
  }
  \label{fig:method_overview}
  \vspace{-4mm}
\end{figure*}

\subsubsection{Single-View Self-Supervision}
For single-view images ($\mathbf{I}_{\mathbf{p}}$), we employ Donor-based Reposing (Eq.~\ref{eq:donor-obj}). We condition on the donor-masked texture $\mathbf{T}_{\mathbf{p} \rightarrow \tilde{\mathbf{p}}}$ and the target pose. Crucially, we explicitly drop the face crop condition ($\mathbf{I}^{FC}_{\text{in}} = \emptyset$) to prevent pixel-perfect information leakage:
\begin{equation}
    \mathbf{c}_{\text{single}} = \{ \mathbf{T}_{\mathbf{p} \rightarrow \tilde{\mathbf{p}}}, \mathbf{p}, \emptyset \}
\end{equation}
This constraint forces the network to recover identity from the warped texture, ensuring robust geometric learning.

In both cases the supervision target (and its rendered target pose) is the BC-standardized image, while the input texture comes from the original image. For paired data these are two different images of the same person, whereas for single images the original and its own BC version form the pair, with the face crop dropped to prevent copy-paste shortcutting.

\subsection{Test-Time Personalization via Few-Shot Adaptation}
\label{sec:finetuning}

\begin{wrapfigure}{r}{0.51\textwidth}
    \centering
    \vspace{-\intextsep}
    \includegraphics[width=\linewidth, keepaspectratio]{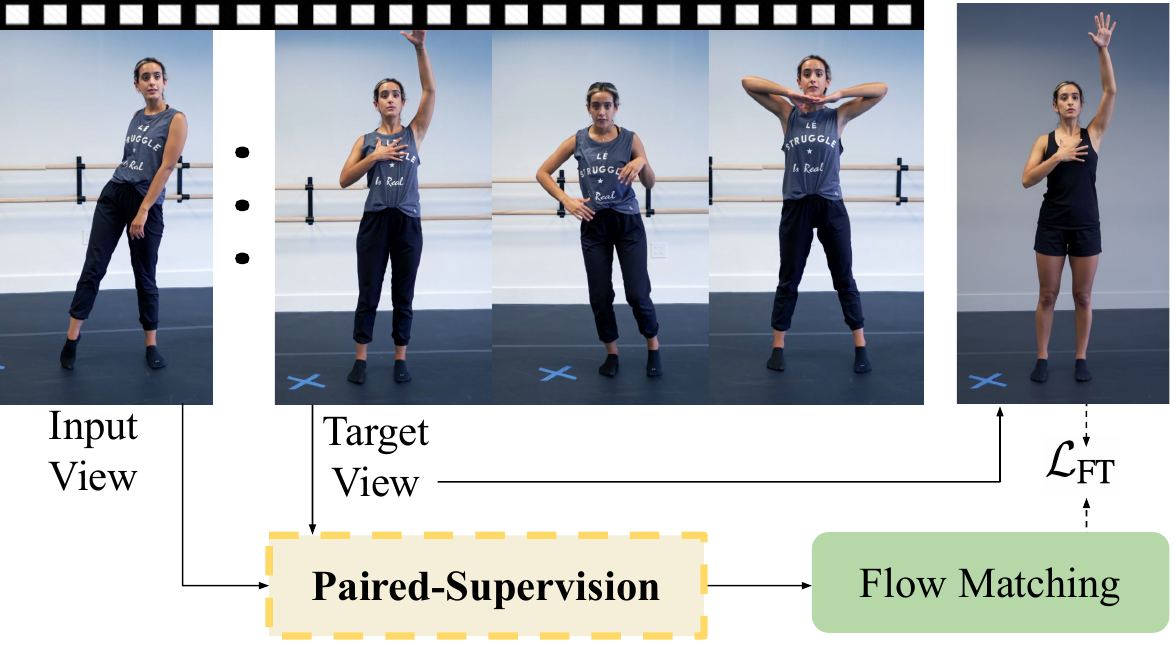}
    \captionof{figure}{\textbf{Finetuning pipeline.} We build input--target pairs from a few-shot subject set and apply a facially masked paired Flow Matching loss to personalize the model at test time.}
    \vspace{-3mm}
    \label{fig:finetuning_method}
\end{wrapfigure}

While our method already achieves high identity preservation in a feed-forward manner, we can further enhance fidelity through test-time personalization via few-shot adaptation. Given a small set $\mathcal{S}$ comprising $N$ images of a person in various poses, we create pairs by using one view as the reference and another as the target. We then fine-tune our LoRA adapters using the paired objective, as illustrated in Figure~\ref{fig:finetuning_method}. This personalizes the model to the specific identity, improving consistency, particularly for extreme pose generation.

We apply the personalization loss to the full visible foreground using a binary mask $\mathbf{M}_j$, which represents the target's latent-space skin segmentation mask:

\begin{flalign}
\label{eq:ft_loss}
\resizebox{0.86\hsize}{!}{$
  \mathcal{L}_{\text{FT}} = \mathbb{E}_{t, (i, j)} \left[ \| \mathbf{M}_j \odot \left( v_{\theta}(\mathbf{x}_{j,t}, t, \mathbf{c}_{\text{paired}}) - v_{\text{target}} \right) \|^2 \right]
$}
\end{flalign}

where $\mathbf{c}_{\text{paired}}=\{ \mathbf{T}_i, \mathbf{p}_j, \mathbf{I}^{FC}_i \}$ is the conditions extracted from a sample pair $(i, j)$ drawn from $\mathcal{S}$ (such that $i \neq j$), and $v_{\text{target}} = \mathbf{x}_{j,1} - \mathbf{x}_{j,0}$ is the target transport guidance.

\subsection{Implementation Details} 
\label{sec:implementation}

\noindent 
\begin{wrapfigure}{r}{0.5\textwidth}
    \centering
    \vspace{-\intextsep}
    \includegraphics[width=\linewidth, keepaspectratio]{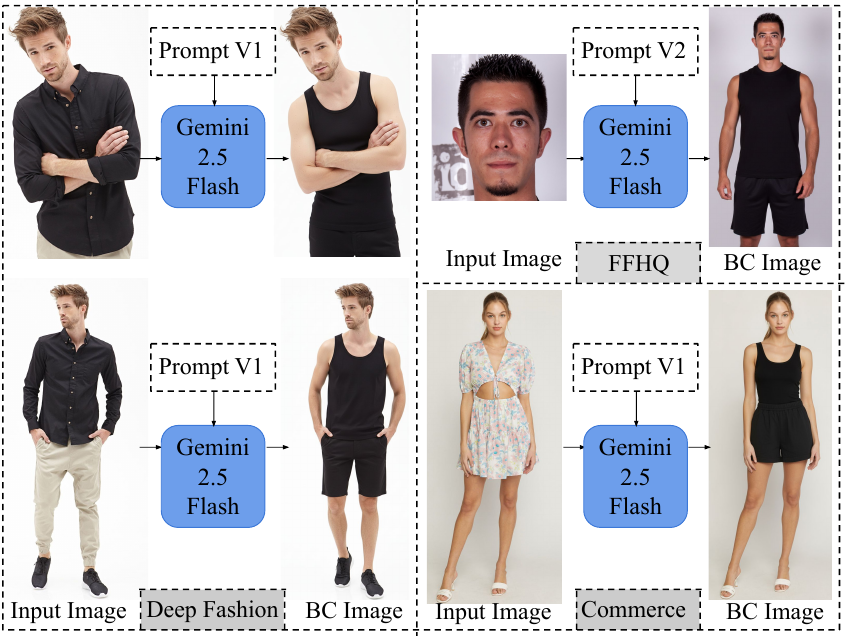}
    \vspace{-3mm}
    \captionof{figure}{\textbf{Base Clothing (BC) Standardization.} We apply different strategies based on the data. For DeepFashion~\cite{deepfashion} and Commerce images~\cite{Zhu2024MMVTO, Cheng2023TryOnDiffusion}, we generate the base garment while preserving pose and identity via pixel-aligned editing (Prompt V1). For FFHQ~\cite{ffhq}, we use generative outpainting to expand limited face crops into full-body samples (Prompt V2).}
    \label{fig:dataset_examples}
    \vspace{-8mm}
\end{wrapfigure}

\textbf{Base Clothing (BC) Dataset.} We construct a standardized dataset by processing paired DeepFashion~\cite{deepfashion} data, unpaired FFHQ~\cite{ffhq} faces, and commercial images~\cite{Zhu2024MMVTO, Cheng2023TryOnDiffusion} using Gemini 2.5 Flash Image~\cite{nanobanana}. As shown in Figure~\ref{fig:dataset_examples}, we employ distinct strategies: \textit{pixel-aligned editing} for full-body inputs to standardize garments to a black tank top and shorts, and \textit{generative outpainting} to expand face crops into full-body portraits. The prompts used for this standardization are provided in the Appendix. While we do not explicitly handle target-pose lighting, BC standardization either re-lights subjects into even studio shading or matches the source illumination; by distilling this data, the model learns to attenuate baked-in highlights from the input texture and render lighting consistent with the input, yielding realistic illumination under novel poses.

\noindent \textbf{Training Setup.} Our dataset totals approximately 500k samples: 470K single unpaired images \cite{ffhq, Zhu2024MMVTO,Cheng2023TryOnDiffusion}, and 30K paired images from the DeepFashion dataset~\cite{deepfashion} (We restrict training to a subset of DeepFashion data to prevent overfitting to its limited number of unique identities ($\approx 100$), while ensuring all identities are still represented.)

\noindent \textbf{Optimization.} We use Flux.1 [dev] ~\cite{flux2024} as our backbone, training rank-128 LoRA adapters with AdamW with a learning rate of $10^{-4}$. Training runs for 100K iterations on 128 TPUv5 chips with a batch size of 128. At inference, our model synthesizes a single reposed avatar in approximately $51$ seconds.

\noindent \textbf{Donor Pool \& Dropout.} For every single-view sample, we pre-compute a pool of 10 valid donor masks $\mathbf{M}_{\tilde{\mathbf{p}}}$ that satisfy an IoU constraint of $[0.4, 0.8]$ with the source mask. To prevent modality over-reliance, we apply mutually exclusive conditioning dropout: we either drop all conditions ($p=0.05$), or individual components: $\mathbf{T}_{\text{in}}$ ($p=0.3$), $\mathbf{I}^{FC}_{\text{in}}$ ($p=0.3$), or $\mathbf{p}_{\text{target}}$ ($p=0.1$).
\vspace{-1mm}
\section{Experiments}
In this section, we lay out the details of our evaluation strategy, including datasets, baselines, and metrics used to compare our method against state-of-the-art approaches. We also conduct extensive ablations to highlight the effectiveness of each of our contributions.

\subsection{Datasets and Metrics for Evaluation}

Following prior work~\cite{MCLD, CFLD, Zhou2025LEFFA}, we use the DeepFashion In-Shop Clothes Retrieval test split~\cite{deepfashion} (8570 image pairs). To assess generalization to in-the-wild scenarios, we additionally evaluate on WPose dataset~\cite{Li_UniHuman_2024} (2305 image pairs).

Since our method canonicalizes clothing, direct comparison against ground-truth images (which include diverse garments and backgrounds) is not meaningful. For image similarity (PSNR, SSIM~\cite{ssim-iqa}) and perceptual metrics (FID~\cite{FID}, LPIPS~\cite{lpips}), we therefore compare our outputs with the BC version. We further report these metrics on the original image only face regions in the appendix.

\textbf{Pose, Identity, Semantic and Perceptual Metrics.} To measure how well the generated avatar matches the target \textbf{pose}, we compute Object Keypoint Similarity (OKS)~\cite{lin2014microsoft} between the predicted and ground-truth keypoints. \textbf{Identity} fidelity is evaluated within the facial region. We compute Face Similarity (FaceSim) using cosine similarity of ArcFace~\cite{arcface} identity embeddings. We report DINOv2 similarity~\cite{oquab2023dinov2} as a measure of \textbf{semantic} and structural alignment. For reference-free \textbf{perceptual} quality, we report HPSv3~\cite{Ma_2025_ICCV}.

\begin{table*}[t!]
\centering

\resizebox{1.0\textwidth}{!}{%
\begin{tabular}{l|cccccccc}
\toprule
\multicolumn{9}{c}{\textbf{DeepFashion (In-Domain)}} \\ \midrule
Method & \textbf{PSNR}$\uparrow$ & \textbf{FID}$\downarrow$ & \textbf{SSIM}$\uparrow$ & \textbf{LPIPS}$\downarrow$ & \textbf{OKS}$\uparrow$ & \textbf{FaceSim}$\uparrow$ & \textbf{DINO}$\uparrow$ & \textbf{HPSv3}$\uparrow$ \\ \hline
CFLD \cite{CFLD} 
& 17.65 & 7.15 & 0.748 & 0.182 & \second{0.48} & 0.3180 & 0.9731 & 4.15 \\
MCLD \cite{MCLD} 
& 18.21 & 7.08 & 0.756 & 0.176 & \best{0.49} & 0.3440 & 0.9654 & 4.29 \\
LEFFA \cite{Zhou2025LEFFA} 
& 14.02 & \second{4.23} & 0.755 & 0.119 & 0.44 & 0.5794 & 0.9409 & 4.41 \\
OnePoseTrans \cite{oneposetrans} 
& 13.57 & 8.74 & 0.605 & 0.307 & 0.46 & 0.5750 & 0.9476 & 4.32 \\
UniHuman \cite{Li_UniHuman_2024} 
& 14.05 & 6.25 & 0.796 & 0.156 & 0.46 & 0.5810 & 0.9434 & 4.03 \\
Gemini 2.5 Flash Image~\cite{nanobanana} 
& 16.98 & 4.59 & 0.738 & 0.179 & 0.43 & 0.5815 & 0.9691 & 7.19 \\ 
Gemini 3 Pro Image~\cite{nanobananapro} 
& 17.51 & 4.30 & 0.775 & 0.109 & 0.45 & 0.5856 & 0.9705 & 7.22 \\ \midrule

Unpaired Only 
& 15.66 & 6.54 & 0.715 & 0.201 & 0.47 & 0.3585 & 0.9259 & 4.25 \\
Paired Only 
& \best{19.38} & \best{4.19} & \second{0.815} & \best{0.071} & \second{0.48} & \best{0.6255} & \best{0.9761} & \best{7.25} \\
\textbf{Ours (Unpaired + paired)} 
& \second{19.36} & 4.24 & \best{0.818} & \second{0.075} & \second{0.48} & \second{0.6047} & \second{0.9759} & \second{7.24} \\  \bottomrule

\multicolumn{9}{c}{\textbf{WPose (Out-of-Domain)}} \\ \midrule
Method & \textbf{M-PSNR}$\uparrow$ & \textbf{FID}$\downarrow$ & \textbf{M-SSIM}$\uparrow$ & \textbf{M-LPIPS}$\downarrow$ & \textbf{OKS}$\uparrow$ & \textbf{FaceSim}$\uparrow$ & \textbf{DINO}$\uparrow$ & \textbf{HPSv3}$\uparrow$ \\ \hline
CFLD \cite{CFLD} 
& 15.43 & 96.07 & 0.744 & 0.208 & 0.31 & 0.0885 & 0.6412 & 1.94 \\
MCLD \cite{MCLD} 
& 15.64 & 94.23 & 0.759 & 0.201 & {0.35} & 0.0995 & 0.6478 & 1.96 \\
LEFFA \cite{Zhou2025LEFFA} 
& 16.71 & 67.85 & 0.776 & 0.193 & 0.32 & 0.0914 & 0.5725 & 2.01 \\
OnePoseTrans \cite{oneposetrans} 
& 17.23 & 27.43 & 0.818 & 0.151 & 0.33 & 0.1735 & 0.7205 & 4.44 \\
UniHuman \cite{Li_UniHuman_2024} 
& 17.64 & 27.75 & 0.807 & 0.161 & 0.34 & 0.1121 & \second{0.7207} & 2.89 \\
Gemini 2.5 Flash Image~\cite{nanobanana} 
& 16.67 & 9.55 & 0.779 & 0.149 & 0.32 & 0.4713 & 0.7005 & 7.35 \\
Gemini 3 Pro Image~\cite{nanobananapro} 
& 17.19 & 7.15 & 0.795 & \second{0.145} & 0.33 & \second{0.5241} & 0.7119 & \second{7.4} \\ \midrule

Unpaired Only 
& 16.13 & 6.95 & 0.761 & 0.215 & \second{0.37} & 0.3805 & 0.6779 & 4.37 \\
Paired Only 
& \second{18.30} & \second{6.65} & \second{0.820} & 0.155 & 0.34 & 0.4959 & 0.6972 & 7.30 \\ 
\textbf{Ours (Unpaired + paired)} 
& \best{19.95} & \best{5.99} & \best{0.860} & \best{0.121} & \best{0.38} & \best{0.5571} & \best{0.7394} & \best{7.55} \\ 
\bottomrule
\end{tabular}%
}
\caption{Quantitative evaluation and ablation study. We report results on DeepFashion (top) and WPose (bottom). The upper section of each block compares our method against state-of-the-art baselines, while the lower section details our ablation study (Unpaired Only, Paired Only, and our full hybrid model). All Pro-Pose results are zero-shot (without test-time fine-tuning). \best{Red} indicates the best performance and \second{Blue} indicates the second best across all methods.}
\label{tab:merged-ablation-quant}

\vspace{-3mm} 

\includegraphics[width=0.95\linewidth]{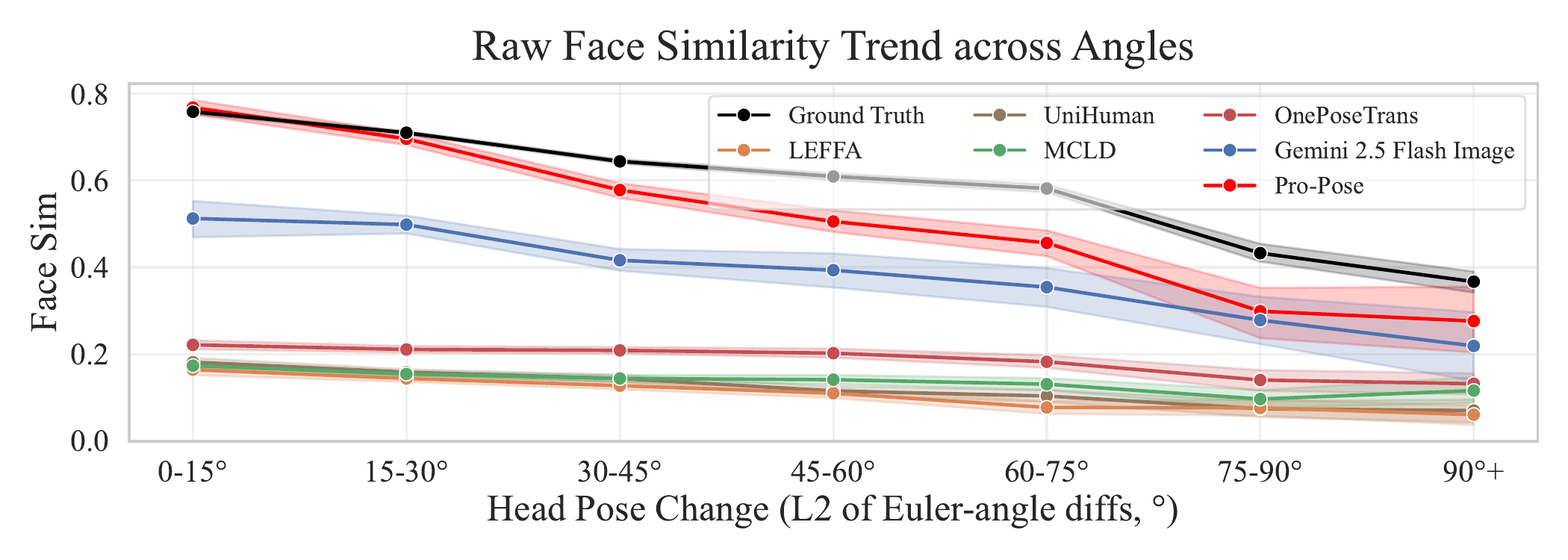}
\vspace{-1mm}
\captionof{figure}{\textbf{Identity Preservation vs.\ Pose Difficulty.} FaceSim scores stratified by pose-change magnitude. Pro-Pose closely tracks Ground Truth identity variance across all difficulty bins, while baselines degrade rapidly. With identity breakage defined as FaceSim $< 0.4$, Pro-Pose fails in only $13.5\%$ of cases vs.\ $> 97\%$ for all baselines.}
\label{fig:stratified_facesim}
\vspace{-8mm}
\end{table*}

\subsection{Quantitative and Qualitative Comparison}
We compare our method with recent diffusion-based pose-conditioned avatar generation models, including MCLD~\cite{MCLD}, CFLD~\cite{CFLD}, LEFFA~\cite{Zhou2025LEFFA}, OnePoseTrans~\cite{oneposetrans}, UniHuman~\cite{Li_UniHuman_2024}, Gemini 2.5 Flash Image (Nano Banana)~\cite{nanobanana}, and Gemini 3 Pro Image (Nano Banana Pro)~\cite{nanobananapro}.

As detailed in Table~\ref{tab:merged-ablation-quant}, our method achieves SOTA performance on both benchmarks. On the in-domain DeepFashion set \cite{deepfashion}, our model sets a new SOTA, significantly outperforming all baselines across image fidelity (PSNR, SSIM, LPIPS), identity (FaceSim), and perceptual quality (HPSv3). For the in-the-wild WPose dataset \cite{Li_UniHuman_2024}, where we use foreground-masked metrics (M-PSNR, M-SSIM, M-LPIPS), our full model (\textit{Unpaired + Paired}) conclusively dominates all SOTA baselines. This highlights the critical role of our unpaired data strategy in achieving robustness for challenging, in-the-wild data.

Figure \ref{fig:qual_all} provides a qualitative comparison with SOTA methods. As demonstrated, our method generates high-fidelity avatars that accurately preserve the input person's identity, facial features, and body characteristics. For instance, the eighth row shows robust preservation of face and beard shape. Furthermore, our approach more robustly follows the target pose than any other method (e.g., sixth row) and maintains body shape fidelity relative to the original image (e.g., ninth row).

\noindent\textbf{Stratified Identity Analysis.} To quantify identity robustness under varying difficulty, we stratify FaceSim scores by the magnitude of pose change between source and target. As shown in Figure~\ref{fig:stratified_facesim}, Pro-Pose closely tracks the inherent identity variance of Ground Truth images across all pose-change bins, whereas all baselines exhibit rapid degradation as pose difficulty increases. Defining identity breakage as a FaceSim score below $0.4$, Pro-Pose exhibits a failure rate of only $13.5\%$, while all competing methods (LEFFA, MCLD, UniHuman, OnePoseTrans) fail to preserve identity in over $97\%$ of generated samples.

\begin{figure*}[t]
  \centering
  \includegraphics[width=0.95\linewidth]{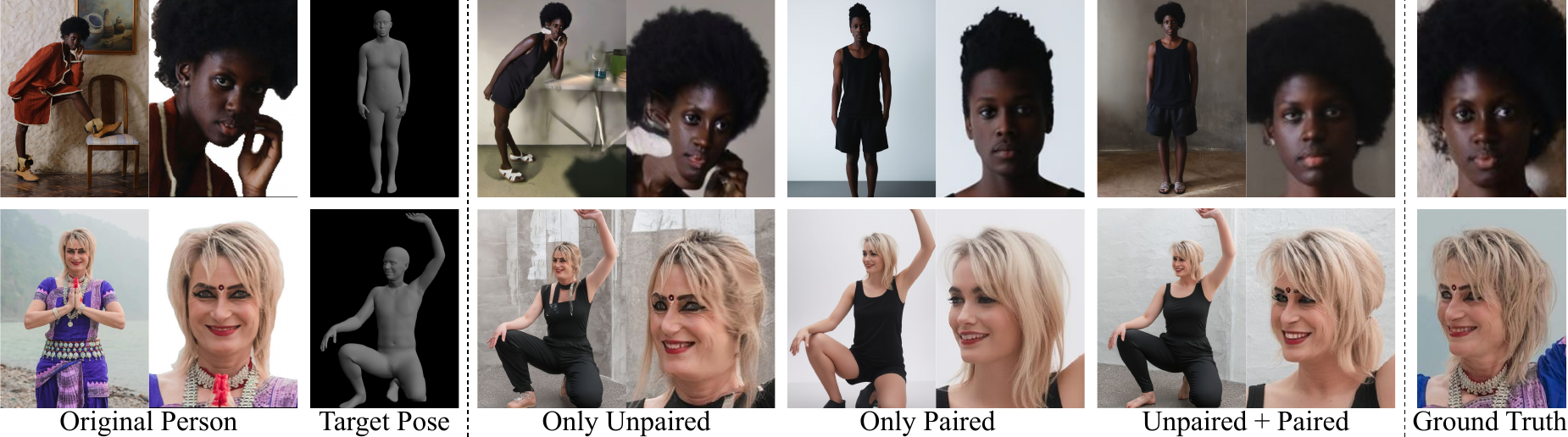}
  \vspace{-2mm}
  \caption{\textbf{Ablation Study of Training Data Sources.}
  We evaluate the impact of different training dataset combinations on generated avatar quality.
  The result columns display models trained on Unpaired data only, Paired data only, and the full Unpaired + Paired combination, respectively. One can observe the improvement in fidelity with the full combined dataset.
  }
  \vspace{-6mm}
  \label{fig:ablation_v1}
\end{figure*}

\vspace{-4mm}
\subsection{Ablation Study}
\label{ssec:ablation}

\textbf{Hybrid Data Strategy.} 
We validate our hybrid data strategy by training three variants of our model: (i) \textit{Unpaired Only}, trained exclusively on unpaired single images; (ii) \textit{Paired Only}, trained exclusively on paired images; and (iii) \textit{Unpaired + Paired}, our full model.

As shown in Table~\ref{tab:merged-ablation-quant}, while the \textit{Paired Only} model performs well on in-domain data (DeepFashion \cite{deepfashion}), it suffers on the out-of-domain WPose set \cite{Li_UniHuman_2024}, confirming its poor generalization. Figure~\ref{fig:ablation_v1} demonstrates that training solely on the limited paired dataset leads to overfitting and identity leaking from the training data. Alternatively, the model trained exclusively on unpaired data fails when subjected to significant pose changes. Our full method (\textit{Unpaired + Paired}) synergizes these sources, effectively utilizing the larger dataset to learn robust identity priors while maintaining stable pose guidance.

\noindent\textbf{Test-Time Personalization.} 

\begin{wrapfigure}{r}{0.48\textwidth}
    \centering
    \vspace{-\intextsep}
    \includegraphics[width=\linewidth]{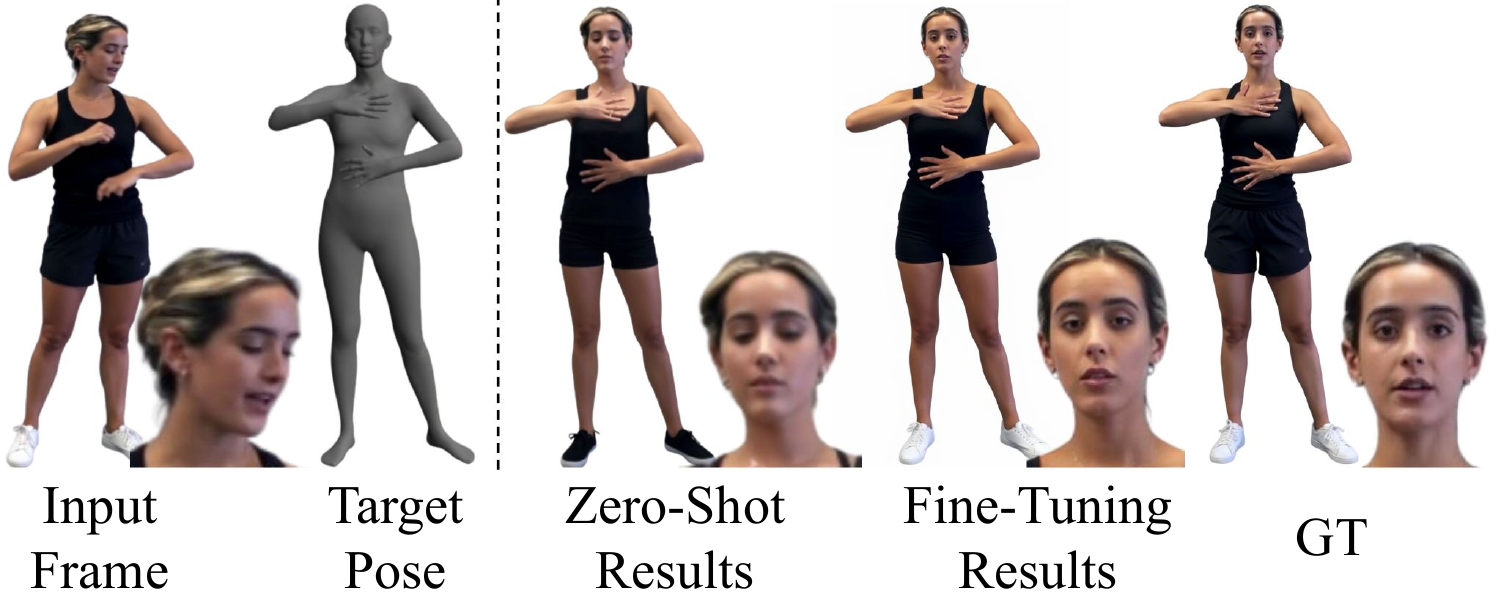}
    \vspace{-6mm}
    \captionof{figure}{\textbf{Finetuning Qualitative results.} Although the input image suffered from motion blur, leading to blurry base model results, finetuning successfully restored crisp face features, demonstrating the model's ability to learn robustly from multiple images rather than relying solely on one input.}
    \label{fig:ft_qual}
    \vspace{1mm}
    \resizebox{\linewidth}{!}{%
    \small
    \begin{tabular}{l|rrr|rrr}
    \hline
    \multicolumn{1}{c|}{Method} &
     \multicolumn{1}{c}{\textbf{PSNR}} &
     \multicolumn{1}{c}{\textbf{FID}} &
     \multicolumn{1}{c|}{\textbf{SSIM}} &
     \multicolumn{1}{c}{\textbf{FaceSim}} &
     \multicolumn{1}{c}{\textbf{DINO}} &
     \multicolumn{1}{c}{\textbf{HPSv3}} \\ \hline
    Ours (Zero-Shot) &
     18.59 &
     6.88 &
     0.823 &
     0.4837 &
     0.689 &
     7.15 \\
    Ours (Fine-Tuned) &
     \textbf{19.34} &
     \textbf{6.76} &
     \textbf{0.835} &
     \textbf{0.5722} &
     \textbf{0.705} &
     \textbf{7.17} \\
     \bottomrule
    \end{tabular}%
    }
    \vspace{-4mm}
    \captionof{table}{\textbf{Ablation study (Fine-Tuning).} Test-time fine-tuning with a few reference images improves identity preservation over our zero-shot base model across all metrics.}
    \label{tab:Ft-table}
    \vspace{-20mm}
\end{wrapfigure}

Furthermore, we validate our test-time personalization strategy, detailed in Section~\ref{sec:finetuning}. As shown in Table~\ref{tab:Ft-table} and qualitatively in Figure~\ref{fig:ft_qual}, applying this few-shot adaptation further improves performance over our base model across all metrics. The most significant gain is in identity preservation, with FaceSim improving by 18.3\%. This confirms that our finetuning approach is highly effective at enhancing model personalization and identity fidelity for specific subjects. We share more finetuning results in the Appendix.

\vspace{-7mm}
\textcolor{black}{\subsection{Downstream Application: Virtual Try-On}
\vspace{-1mm}
\label{ssec:vto_application}
A key advantage of our canonical representation is its utility as a preprocessing stage for downstream tasks. We demonstrate this by combining Pro-Pose with the state-of-the-art Google Virtual Try-On tool~\cite{googlevto2023}. As shown in Figure~\ref{fig:vto_application}, applying VTO directly to in-the-wild images can lead to erroneous results due to challenging poses, cluttered backgrounds, and complex clothing. In contrast, first generating a clean, reposed canonical avatar with Pro-Pose significantly improves VTO fidelity.}

\begin{figure}[ht]
\centering
\includegraphics[width=0.95\linewidth]{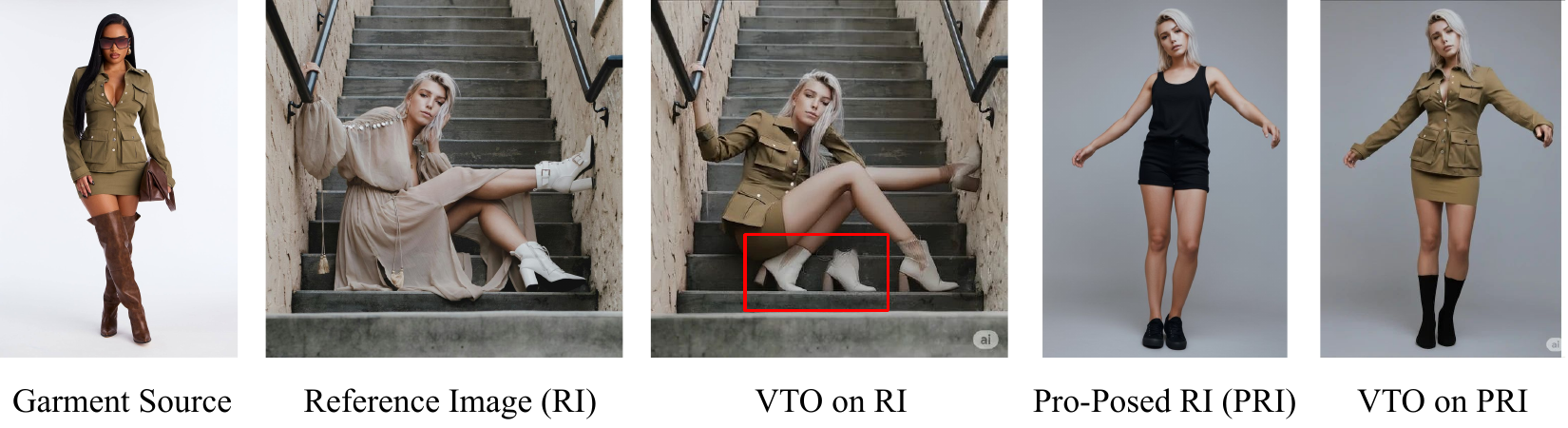}
\caption{\textbf{Downstream Application: Virtual Try-On.} Applying VTO~\cite{googlevto2023} directly to an unconstrained in-the-wild image (VTO on RI) yields limited fidelity. Using Pro-Pose to first generate a clean, reposed canonical avatar (Pro-Posed RI) significantly improves VTO quality (VTO on PRI).}
\label{fig:vto_application}
\end{figure}

\begin{wraptable}{r}{0.46\textwidth}
\centering
\small
\setlength{\tabcolsep}{4pt}
\begin{tabular}{l|cc}
\toprule
\textbf{Error / Method} & \textbf{Orig.} & \textbf{Pro-Pose} \\ \midrule
VTO pref. $\uparrow$ & 28.02\% & \textbf{71.98\%} \\
Garment err. $\downarrow$ & 31.04\% & \textbf{17.34\%} \\
Person err. $\downarrow$ & \textbf{2.51\%} & 3.02\% \\
\bottomrule
\end{tabular}
\caption{\textbf{VTO User Study.} VTO on Pro-Posed avatars is strongly preferred over VTO on originals under a matched pose. The slight person-error uptick is attributed to SMPL-X limitations (Sec.~\ref{sec:limits}).}
\label{tab:user-study}
\vspace{-\intextsep}
\end{wraptable}
\noindent To fairly quantify this benefit, we conducted a user study on paired WPose samples in which both branches target the same pose: Branch-A applies VTO directly to the original image, while Branch-B applies VTO to the Pro-Posed avatar. Across 30K votes (10 raters over 3K pairs; 500 images $\times$ 6 garments), raters preferred VTO on Pro-Posed avatars in $71.98\%$ of comparisons, with substantially fewer garment errors (Table~\ref{tab:user-study}).

\subsection{Emergent Model Properties: Direct Texture Editing.}

\begin{wrapfigure}{r}{0.5\textwidth}
    \centering
    \vspace{-\intextsep}
    \includegraphics[width=\linewidth]{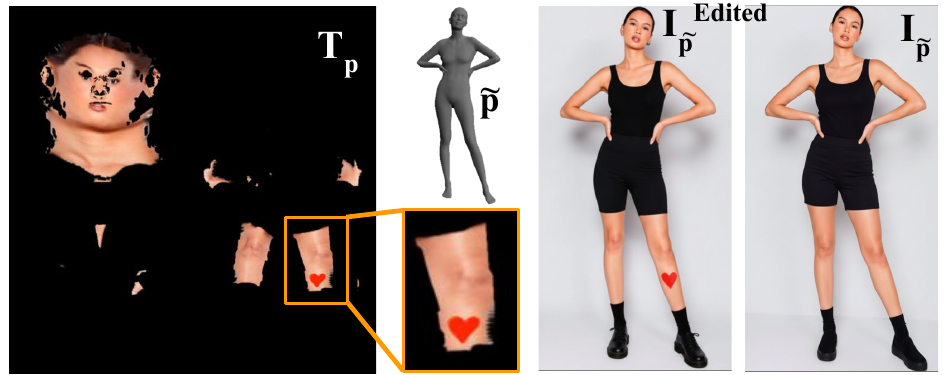}
    \vspace{-1.5em}
    \captionof{figure}{\textbf{Emergent Capabilities.} \textbf{$T_p$}: Edited texture map. By introducing a localized modification — in this case, adding a heart tattoo to the base texture — the edit geometrically propagates with high fidelity to novel target poses $\tilde{p}$ as shown in $I^{Edited}_{\tilde{p}}$. $I_{\tilde{p}}$ shows the output using the un-edited texture map. This demonstrate our model's capability to apply texture changed such as make up or tattoo in synthesizing the output portraits.}
    \label{fig:emergent}
    \vspace{-16mm}
\end{wrapfigure}

Our framework unlocks other powerful features without explicit training such as direct texture editing.  Because Pro-Pose extracts a canonical UV map, any manual edits to this texture (e.g., adding a tattoo) propagate with strict geometric consistency to the final reposed avatar, preserving identity without distortion (Figure~\ref{fig:emergent} adding a heart tattoo on texture map can transfer to the reposed person).

\begin{wrapfigure}{r}{0.53\textwidth}
    \centering
    \vspace{-26mm}
    \includegraphics[width=\linewidth]{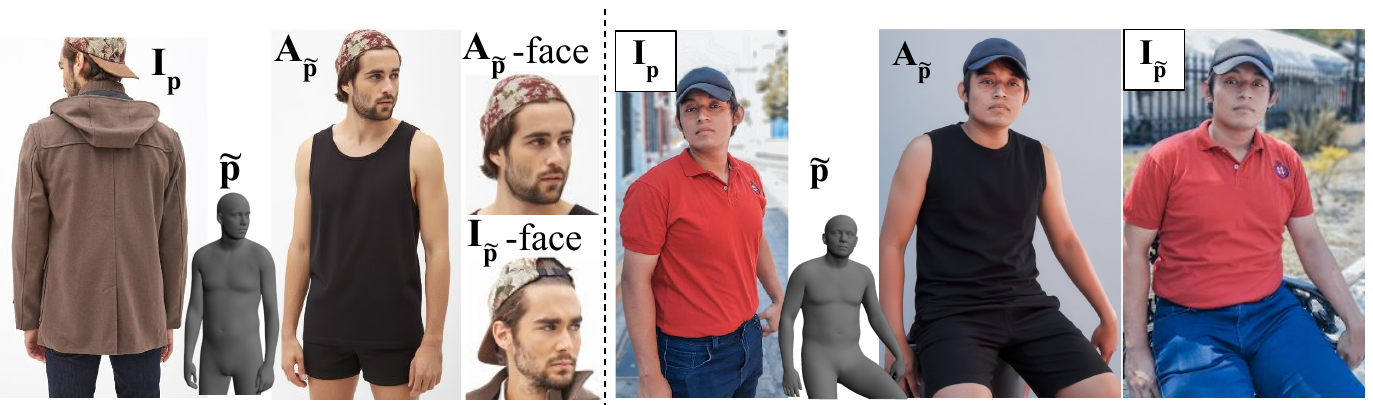}
    \vspace{-5mm}
    \caption{\textbf{Limitations of the base model.} \textbf{Left:} Extreme pose changes (e.g., back-to-front) result in minimal UV texture overlap, forcing the model to hallucinate unseen regions. While general attributes like beard shape or hair color may be inferred from partial cues, precise facial structure can lose fidelity. \textbf{Right:} Zero-shot synthesis is constrained by the accuracy of single-view 3D body estimation. Errors in SMPL-X fitting can propagate during the reposing stage, leading to inaccurate body proportions (e.g., a noticeable reduction in thigh volume in the ProPose results compared to the ground truth).}
    \vspace{-6mm}
    \label{fig:limitations}
\end{wrapfigure}
\section{Limitations and Future Work}
\label{sec:limits}

While our framework significantly advances single-image avatar reposing, some limitations remain. First, the base model struggles with extreme pose changes, such as synthesizing a frontal pose from a single rear-view image. Consequently, the model inevitably hallucinates facial features (Figure~\ref{fig:limitations}, left). However, this limitation can be mitigated by providing multiple reference images of the subject and applying our fine-tuning approach. Second, our method relies heavily on the accuracy of the underlying SMPL-X body estimation. During zero-shot in-the-wild evaluation, we extract body shape parameters exclusively from the single reference image and combine them with the pose parameters of the target. However, recovering 3D body shape from a single image can be erroneous. Figure~\ref{fig:limitations} (right) showcases an example from the WPose dataset where the reference body shape is inaccurately estimated. When this flawed shape geometry is reposed into a novel target view, the errors propagates directly to the generated canonical avatar and resulting in incorrect body shape. This highlights that our synthesis fidelity is ultimately bounded by the robustness of single-view 3D body priors. To mitigate such failures in practice, we leverage the pose-conditioning dropout used during training to drop unreliable pose estimates, or replace them with the $\hat{\beta}$ of a similar-bodied reference subject; this reduces gross geometric artifacts at the cost of exact body-shape fidelity. Third, while our full-body UV conditioning preserves visible skin details, self-occluded body regions cannot be faithfully reconstructed from a single view. Future work will explore more robust body estimation techniques and extend the framework to zero-shot multi-reference consistency.
\section{Conclusion}
\label{sec:conclusion}

In this paper, we presented a novel framework for generating photorealistic, reposed human avatars from unconstrained single images. By introducing a self-supervised \textit{Donor-based UV Reposing} strategy operating on full-body UV textures, we effectively decoupled pose from appearance, enabling our Flow Matching generator to learn robust full-body identity preservation from massive unpaired image collections. We further showed that this strong base model can be rapidly specialized via multi-image fine-tuning to handle more extreme reposing scenarios. By standardizing data into a base clothing and formulating generation in canonical UV space, our approach provides a scalable foundation for creating high-fidelity, controllable digital avatars from everyday photography, and serves as a powerful canonical intermediate for downstream applications such as virtual try-on.

\begin{figure*}[!ht]
\centering
\includegraphics[width=\linewidth]{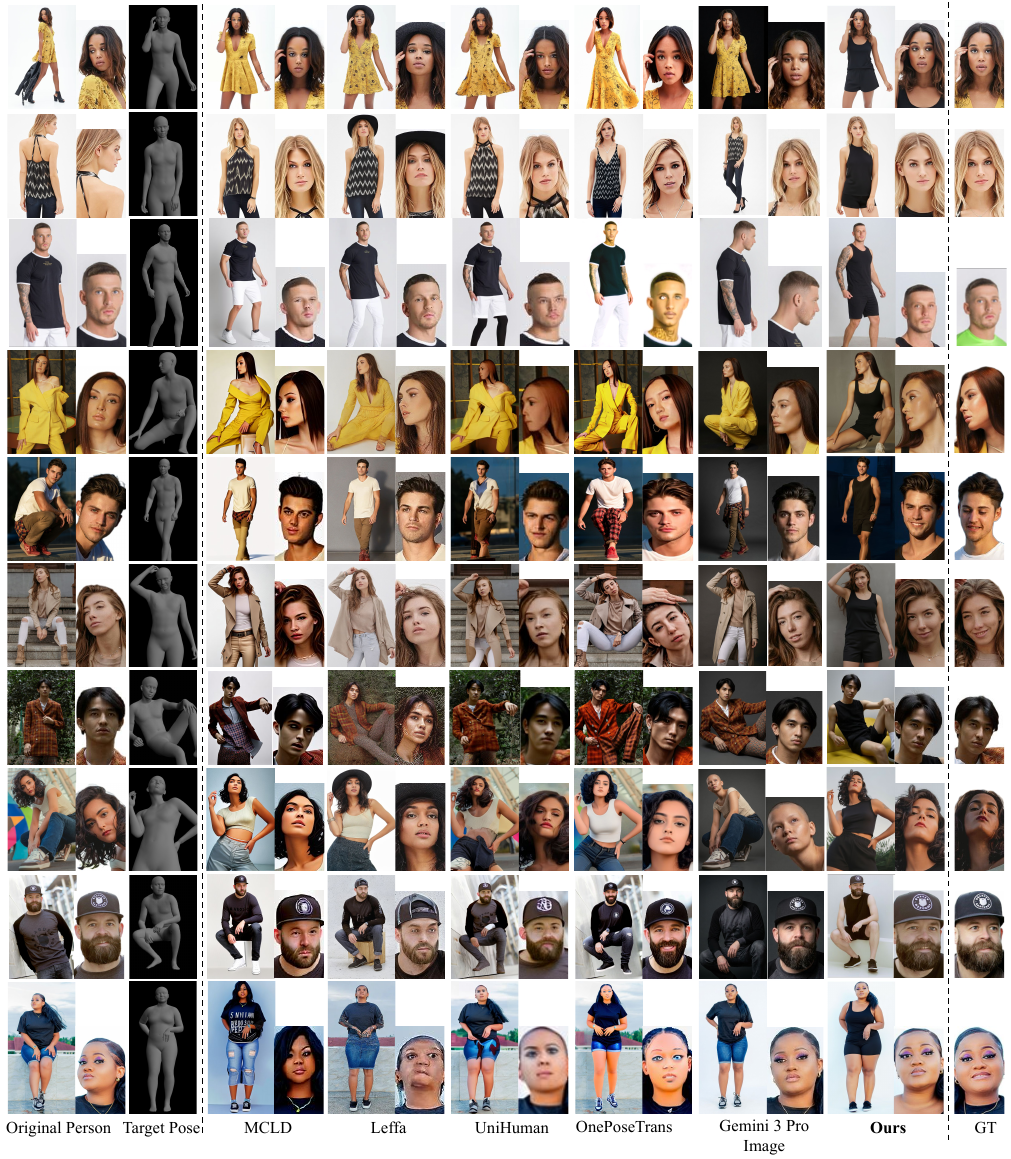}
\vspace{-7mm}
\caption{\textbf{Qualitative Comparison.} Results are shown for the DeepFashion \cite{deepfashion} (rows 1-3) and WPose \cite{Li_UniHuman_2024} (bottom rows) datasets. The columns show the Original Person, Target Pose, and results from state-of-the-art methods (MCLD \cite{MCLD}, Leffa \cite{Zhou2025LEFFA}, UniHuman \cite{Li_UniHuman_2024}, OnePoseTrans \cite{oneposetrans}, Gemini 3 Pro Image~\cite{nanobananapro}) compared to \textbf{Ours (Zero-Shot)} and the Ground Truth (GT). Our method demonstrates superior performance in preserving the person's identity, particularly the facial characteristics, across varying poses. For instance, in the third and fifth rows, previous methods introduce significant identity distortion and loss of facial likeness (e.g., changes in jawline or features). In contrast, \textbf{Ours} robustly preserves the unique face structure and identity of the original person, closely matching the GT result. (prompt used for Gemini 3 Pro Image: {\ttfamily A professional studio portrait of the person from IMAGE\_1, maintaining their exact facial features, hairstyle, build, and identity. They are captured in the pose and camera angle shown in IMAGE\_2}).}
 \label{fig:qual_all}
 \vspace{-10mm}
\end{figure*}
\newpage
\bibliographystyle{splncs04}
\bibliography{main}

\newpage
\appendix
\begin{center}
{\LARGE\bfseries Supplementary Material}
\end{center}
\vspace{1cm}
\suppressfloats[t]
\vspace{-16mm}
\section{Base Clothing (BC) Dataset.}
To train our model, we utilize paired data from DeepFashion~\cite{deepfashion}, unpaired single images from FFHQ~\cite{ffhq}, and commercial data derived from e-commerce websites~\cite{Zhu2024MMVTO, Cheng2023TryOnDiffusion}. To ensure visual uniformity across these diverse sources, we generate the \textbf{Base Clothing (BC)} dataset using an identity-preserving I2I model (Gemini 2.5 Flash Image~\cite{nanobanana}), as shown in Figure 5 (main paper). Depending on the nature of the input data, we employ either Prompt V1 or Prompt V2:

\smallskip\noindent\textbf{Prompt V1:}
\begin{quote}\footnotesize\ttfamily\raggedright
Precisely edit IMAGE\_0 to change the garment worn by the person in the photo to a black sleeveless tank top and 3-inch black shorts. Do not change the person in the image or the framing. Remove or replace any accessories such as bags, hats, jewelry and sunglasses in IMAGE\_0. Do not alter IMAGE\_0 in any other way. Keep it pixel-aligned.
\end{quote}

\noindent\textbf{Prompt V2:}
\begin{quote}\footnotesize\ttfamily\raggedright
Precisely edit IMAGE\_0 to generate a full body image of this person (head to toes) while keeping the face and head pose same in a black sleeveless tank top and 3-inch black shorts. Generate the body pose suiting the existing face pose. The person is posing naturally against a clean and plain background. Even, soft studio lighting illuminates the subject from the front, highlighting the form and texture of the garments. No accessories (no bags, hats, jewelry, sunglasses). Professional product photography style, no harsh shadows or distracting elements. High resolution, sharp focus.
\end{quote}

\section{Canonical UV-Space Formulation}
\label{sec:method_formulation}

To formulate our generative objective in the canonical UV space, we first extract the parametric body model and the corresponding partial texture map $\mathbf{T}_\mathbf{p}$ from the input image $\mathbf{I}$.

\noindent\textbf{Pose and Shape Estimation.} 
We utilize the SMPL-X body model \cite{Pavlakos2019SMPLX} to represent the human subject. Given an input image $\mathbf{I}$, we estimate the shape ${\hat{\beta}}$, pose ${\hat{\theta}}$, and expression ${\hat{\psi}}$ parameters via SMPLify-X optimization~\cite{Pavlakos2019SMPLX}. The pose parameters ${\hat{\theta}}$ are the control variable for the target pose. The SMPL-X function $M({\hat{\beta}}, \hat{{\theta}}, \hat{{\psi}})$ maps these parameters to a triangular mesh $\mathcal{M} = (\mathbf{V}, \mathcal{F})$, where $\mathbf{V} \in \mathbb{R}^{N \times 3}$ represents the vertices and $\mathcal{F}$ the fixed topology faces. We obtain the target pose condition $\mathbf{p}$ by rendering the estimated mesh $\mathcal{M}$ into the image space.

\noindent\textbf{Partial Texture Extraction.} 
To obtain the partial texture $\mathbf{T}_p$, we perform an inverse rendering operation via barycentric sampling in UV space. Let $\mathbf{V}^{uv} \in \mathbb{R}^{N \times 2}$ denote the fixed UV coordinates of the SMPL-X topology, and $\mathbf{V}^{img} \in \mathbb{R}^{N \times 2}$ denote the projection of the 3D vertices $\mathbf{V}$ onto the coordinate space of input image $\mathbf{I}$.

Computing $\mathbf{T}_\mathbf{p}$ directly via ray-casting is computationally expensive. Instead, we adopt an efficient rasterization-based approach. We rasterize the mesh in UV space, utilizing $\mathbf{V}^{uv}$ as the position attributes. During rasterization, we interpolate the image-space coordinates $\mathbf{V}^{img}$ across the triangle faces. For a given texel $\mathbf{u}$ in the canonical UV space $\Omega_{uv}$, located within a triangle $f \in \mathcal{F}$ with barycentric weights $\mathbf{b} = (b_1, b_2, b_3)$, the corresponding source image coordinate $\mathbf{x}_{src}$ is computed as:
\begin{equation}
    \mathbf{x}_{src}(\mathbf{u}) = \sum_{k=1}^{3} b_k \cdot \mathbf{V}^{img}_{f,k}
\end{equation}
where $\mathbf{V}^{img}_{f,k}$ corresponds to the image-space coordinate of the $k$-th vertex of face $f$. The pixel value for the partial texture at location $\mathbf{u}$ is then obtained by bilinear sampling of the input image:
\begin{equation}
    \mathbf{T}_p(\mathbf{u}) = 
    \begin{cases} 
        \mathbf{I}(\mathbf{x}_{src}(\mathbf{u})) & \text{if } \mathbf{M}_\mathbf{p}(\mathbf{u}) = 1 \\
        0 & \text{otherwise}
    \end{cases}
\end{equation}
Here, $\mathbf{M}_\mathbf{p} \in \{0,1\}^{H \times W}$ is the binary visibility mask determined by the z-buffer during the initial estimation. To ensure high-fidelity unwrapping and avoid artifacts at UV seams, we filter triangles where the edge lengths in UV space exceed a threshold $\tau_{dist}$.

\paragraph{Design Choice: Partial UVs vs. Generative UV Recovery.}
An alternative to our barycentric partial-texture extraction is generative UV recovery, such as FreeUV~\cite{yang2025freeuv}. FreeUV, however, operates only on the facial (FLAME) region and recovers texture through generative completion. Adopting such a model upstream risks altering the subject's facial identity, and is in any case restricted to the face, whereas our pipeline requires full-body textures. We therefore retain the original partial UV maps, despite their incompleteness, to prevent generative identity loss and preserve whole-body skin appearance, and leave the joint generation of a complete UV texture and the final avatar to future work.

\section{Ablation Study: Donor-based UV Reposing vs Random Patch Masking}
\label{sec:ablation_masking}

In Section 3.2 (main paper), we introduced \textit{Donor-based UV Reposing} to prevent the model from learning shortcuts. We argued that using a real mask from a "donor" creates realistic occlusions that force the network to learn geometry. In this section, we test if this is actually necessary. We compare our method against a simpler baseline where we just block out random parts of the texture to create a synthetic pair.

\noindent \textbf{Implementation Details.} 
For the "Random masking" baseline, we take the input texture ($\mathbf{T}_\mathbf{p}$) and cover it with random black squares. We use up to 6 non-overlapping patches, each $64 \times 64$ pixels. All other training settings remain identical to our main method.

\noindent \textbf{Qualitative Results.} 

\begin{wrapfigure}{r}{0.64\textwidth}
    \centering
    \vspace{-\intextsep}
    \includegraphics[width=\linewidth]{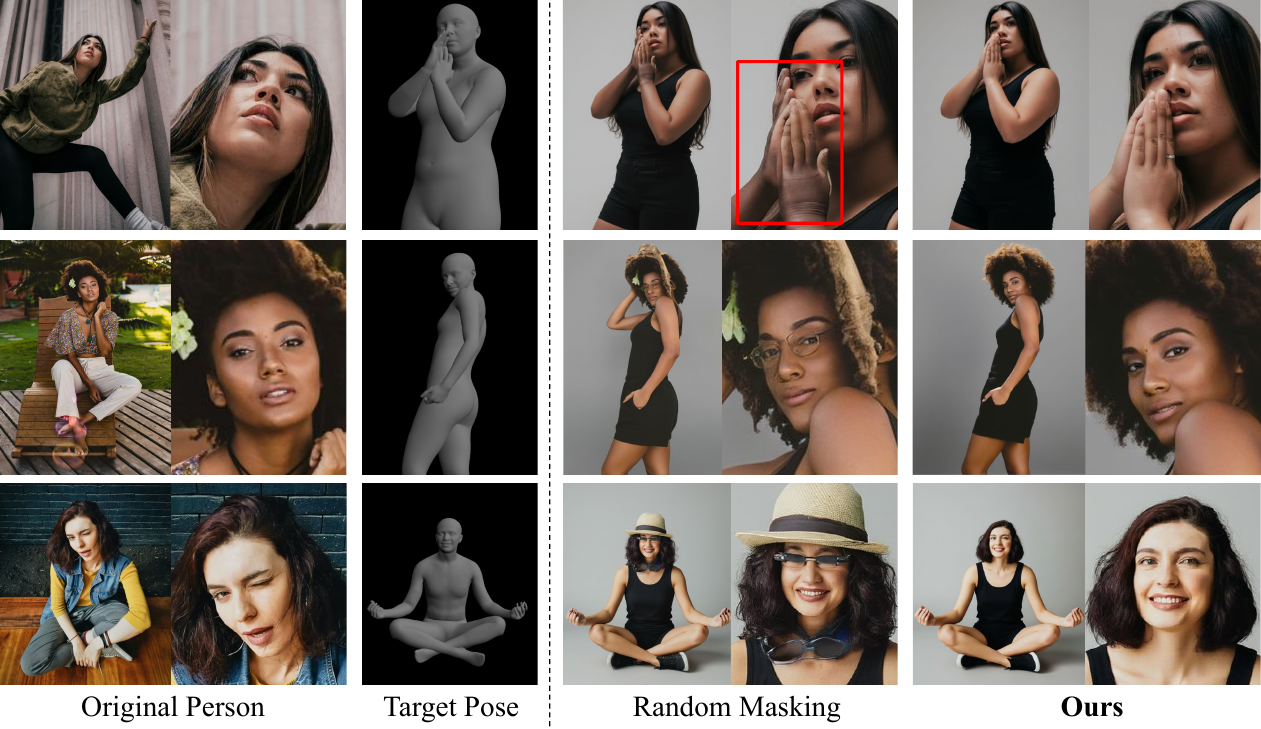}
    \caption{\textbf{Random Patch Masking vs. Ours (Donor-Based UV Reposing).} Random masking preserves UV boundary shape, leaking pose information and causing artifacts (e.g., hallucinated accessories). Our donor-based reposing changes texture boundaries completely, forcing the model to learn geometry rather than shortcuts.}
    \label{fig:masking_ablation_qual}
    \vspace{-5mm}
\end{wrapfigure}

Visual comparisons in Figure~\ref{fig:masking_ablation_qual} show that random masking is insufficient. The problem is that random squares do not change the overall "shape" of the visible texture. The boundaries of the valid texture still look like the source pose.
Because the model can still recognize the source pose from these boundaries, it tries to find a shortcut. However, the random black squares create confusing holes that don't match natural occlusions. To explain these unnatural missing regions, the model, more frequently, hallucinates accessories.

Our Donor-based method avoids this because the donor mask looks like a real human pose. It changes the texture boundaries completely, so the model cannot rely on the visible pose outline and must learn beyond simple "copy-pasting".

\begin{wrapfigure}{r}{0.6\textwidth}
    \centering
    \vspace{-\intextsep}
    \vspace{1mm}
    \resizebox{\linewidth}{!}{%
    \small
    \begin{tabular}{l|cccccc}
    \hline
    Masking & \textbf{PSNR}$\uparrow$ & \textbf{FID}$\downarrow$ & \textbf{SSIM}$\uparrow$ & \textbf{FaceSim}$\uparrow$ & \textbf{DINO}$\uparrow$ & \textbf{HPSv3}$\uparrow$ \\ \hline
    Random & 18.56 & 6.45 & 0.821 & 0.5187 & 0.6965 & 7.35 \\
    \textbf{Ours} & \textbf{19.95} & \textbf{5.99} & \textbf{0.860} & \textbf{0.5571} & \textbf{0.7394} & \textbf{7.55} \\
    \bottomrule
    \end{tabular}%
    }
    \vspace{-2mm}
    \captionof{table}{\textbf{Donor Masking Ablation.} Random masking ($64\times64$ patches) vs.\ our donor-based reposing. Our method outperforms on all metrics, confirming that structurally semantic masks are essential.}
    \label{tab:donor_masking_ablation_quant}
    \vspace{-\intextsep}
    \vspace{1mm}
\end{wrapfigure}

\noindent \textbf{Quantitative Results.} 
Table~\ref{tab:donor_masking_ablation_quant} confirms these visual findings. The Random Masking baseline performs worse on all metrics when evaluated on the WPose dataset. Most notably, identity preservation drops (FaceSim decreases by 6.75\%) and overall image quality degrades. This proves that simply hiding pixels is not enough; the occlusion must mimic a realistic pose to train the model effectively.

\section{Additional Results}
\label{sec:additional_results}

\begin{table}[!t]
\centering
\resizebox{1.0\textwidth}{!}{%
\begin{tabular}{l|cccccccc}
\toprule
\multicolumn{9}{c}{\textbf{DeepFashion (In-Domain) -- BC Inputs}} \\ \midrule
Method & \textbf{PSNR}$\uparrow$ & \textbf{FID}$\downarrow$ & \textbf{SSIM}$\uparrow$ & \textbf{LPIPS}$\downarrow$ & \textbf{OKS}$\uparrow$ & \textbf{FaceSim}$\uparrow$ & \textbf{DINO}$\uparrow$ & \textbf{HPSv3}$\uparrow$ \\ \hline
CFLD \cite{CFLD} 
& 18.51 & 7.09 & 0.757 & 0.180 & {0.48} & 0.3182 & 0.9733 & 4.21 \\
MCLD \cite{MCLD} 
& 19.18 & 7.01 & 0.769 & 0.173 & \textbf{{0.49}} & 0.3437 & 0.9657 & 4.35 \\
LEFFA \cite{Zhou2025LEFFA} 
& 14.95 & \textbf{4.20} & 0.767 & 0.117 & 0.45 & 0.5781 & 0.9415 & 4.40 \\
OnePoseTrans \cite{oneposetrans} 
& 14.35 & 8.59 & 0.618 & 0.287 & 0.46 & 0.5761 & 0.9481 & 4.37 \\
UniHuman \cite{Li_UniHuman_2024} 
& 15.95 & 6.16 & 0.803 & 0.149 & 0.46 & 0.5805 & 0.9436 & 4.24 \\
Gemini 2.5 Flash Image~\cite{nanobanana} 
& 17.59 & 4.44 & 0.757 & 0.167 & 0.43 & 0.5817 & 0.9697 & 7.22 \\ \midrule
\textbf{Ours} 
& \textbf{{19.36}} & 4.24 & \textbf{{0.818}} & \textbf{{0.075}} & {{0.48}} & \textbf{{0.6047}} & \textbf{{0.9759}} & \textbf{7.24} \\  \bottomrule

\multicolumn{9}{c}{\textbf{WPose (Out-of-Domain) -- BC Inputs}} \\ \midrule
Method & \textbf{M-PSNR}$\uparrow$ & \textbf{FID}$\downarrow$ & \textbf{M-SSIM}$\uparrow$ & \textbf{M-LPIPS}$\downarrow$ & \textbf{OKS}$\uparrow$ & \textbf{FaceSim}$\uparrow$ & \textbf{DINO}$\uparrow$ & \textbf{HPSv3}$\uparrow$ \\ \hline
CFLD \cite{CFLD} 
& 15.61 & 91.67 & 0.753 & 0.201 & 0.31 & 0.0888 & 0.6418 & 1.92 \\
MCLD \cite{MCLD} 
& 15.80 & 88.76 & 0.764 & 0.196 & {0.35} & 0.0997 & 0.6479 & 1.93 \\
LEFFA \cite{Zhou2025LEFFA} 
& 16.82 & 64.44 & 0.786 & 0.186 & 0.33 & 0.0911 & 0.5734 & 2.04 \\
OnePoseTrans \cite{oneposetrans} 
& 17.35 & 25.87 & 0.825 & 0.145 & 0.33 & 0.1734 & 0.7214 & 4.39 \\
UniHuman \cite{Li_UniHuman_2024} 
& 17.75 & 24.13 & 0.815 & 0.153 & 0.34 & 0.1121 & {0.7211} & 2.95 \\
Gemini 2.5 Flash Image~\cite{nanobanana} 
& 16.74 & 8.18 & 0.788 & {0.144} & 0.32 & 0.4727 & 0.7021 & {7.38} \\ \midrule
\textbf{Ours} 
& \textbf{{19.95}} & \textbf{{5.99}} & \textbf{{0.860}} &\textbf{ {0.121} }& \textbf{{0.38}} & \textbf{{0.5571}} & \textbf{{0.7394}} & \textbf{7.55} \\ 
\bottomrule
\end{tabular}%
}
\vspace{1mm}
\caption{\textbf{Quantitative Comparison with BC Inputs.} All baselines receive BC-preprocessed input images at test time, removing garment variation as a confounding factor. Pro-Pose maintains its advantage, confirming that our gains are methodological.}
\label{tab:merged-ablation-quant-suppl}

\vspace{-8mm} 
\end{table}

\begin{figure}[t]
    \centering
    \includegraphics[width=\linewidth]{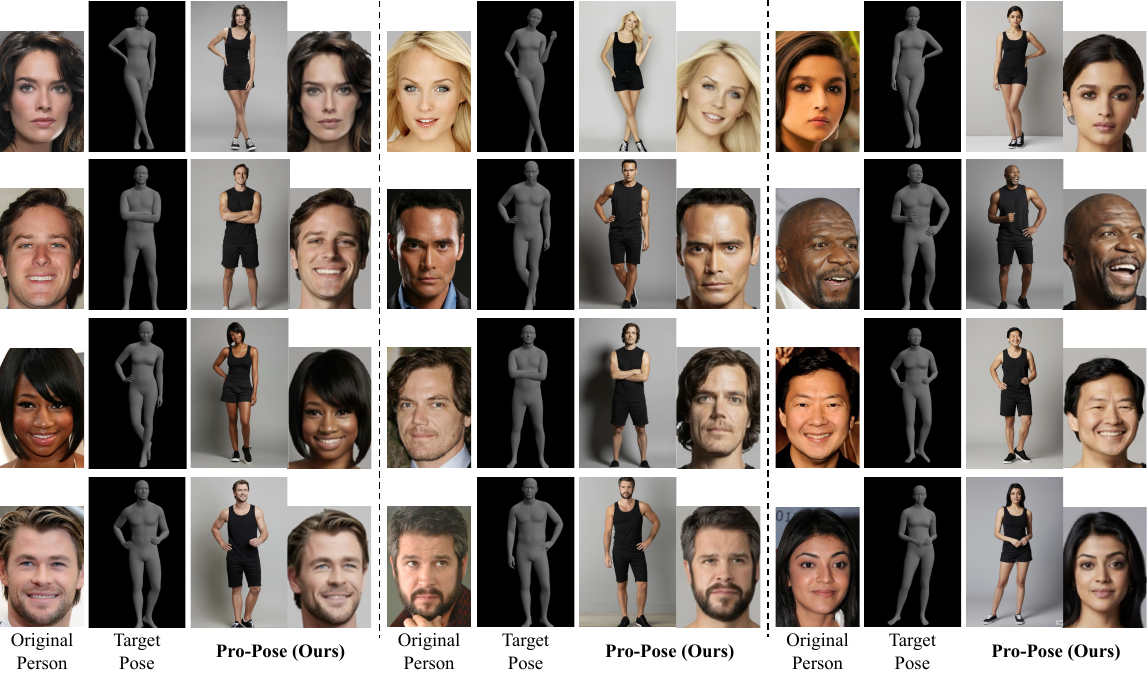}
    \caption{\textbf{Pro-Pose In-the-wild.} We apply our method to real-world images characterized by varying poses and environments. The columns display (from left to right): (1) the original input image; (2) the target pose represented by a SMPL-X mesh (reposed using ground truth parameters); (3) the output generated by Pro-Pose with a corresponding face crop for detailed visual comparison. Our method consistently preserves identity, including facial details, body shape, and visible skin features, compared to the ground truth.}
    \label{fig:qualitative_in_the_wild}
    \vspace{-5mm}
\end{figure}

\subsection{Extended Evaluation In-The-Wild}
\label{ssec:in_the_wild}

To demonstrate the robustness of our approach, we evaluate Pro-Pose on ``in-the-wild'' images sourced from CelebA-HQ~\cite{celebAHQ}. These samples feature diverse lighting conditions, backgrounds, and subjects. The qualitative results, presented in Figure~\ref{fig:qualitative_in_the_wild}, confirm that our method generalizes effectively to unseen real-world data, maintaining high identity fidelity and pose accuracy.

\begin{figure}[htbp!]
  \centering
  \includegraphics[width=1\linewidth]{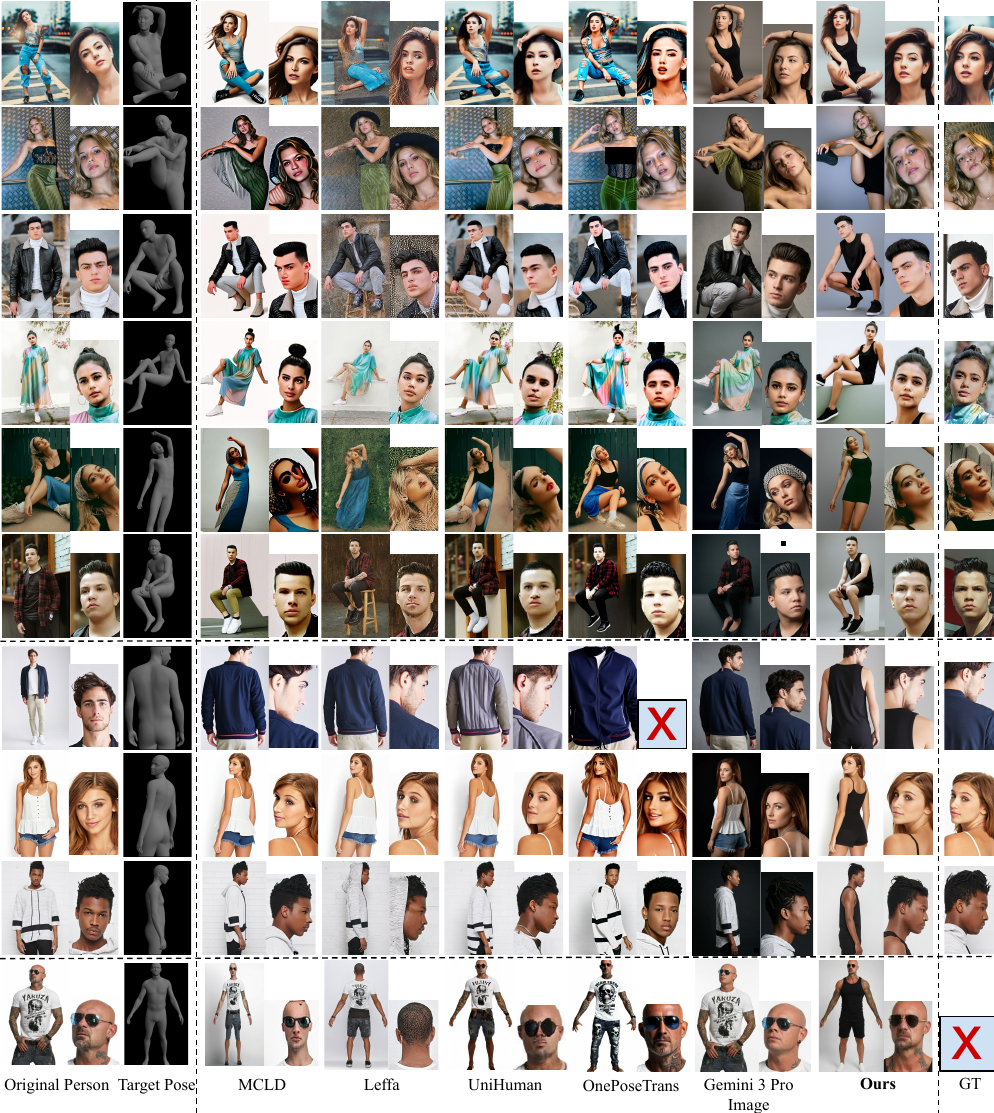}
  \caption{\textbf{Qualitative Comparison.} The last three rows display results from the DeepFashion dataset \cite{deepfashion}, and all previous rows utilize real images from the WPose dataset \cite{Li_UniHuman_2024}. The columns show the Original Person, Target Pose, and results from state-of-the-art methods (MCLD \cite{MCLD}, Leffa \cite{Zhou2025LEFFA}, UniHuman \cite{Li_UniHuman_2024}, OnePoseTrans \cite{oneposetrans}, Gemini 3 Pro \cite{nanobananapro}) compared to \textbf{Ours} and the Ground Truth (GT). Our method demonstrates superior performance in preserving the person's identity, including facial features, body shape, and visible skin characteristics, across varying poses. For instance, in the third and fifth rows, previous methods introduce significant identity distortion and loss of facial likeness (e.g., changes in jawline or features). In contrast, \textbf{Ours} robustly preserves the unique face structure and identity of the original person, closely matching the GT result. For the final row, no ground truth (GT) exists because the input is a single image reposed into a target A-pose. Although a GT face crop cannot be provided, comparing the result to the face crop of the original image demonstrates our strong identity preservation.}
  \label{fig:qual_comparison_appendix}
\end{figure}

\vspace{-2mm}
\subsection{Extended Qualitative Comparisons}
We provide more extensive qualitative comparisons against state-of-the-art methods (MCLD \cite{MCLD}, Leffa \cite{Zhou2025LEFFA}, UniHuman \cite{Li_UniHuman_2024}, OnePoseTrans \cite{oneposetrans}, and Gemini 2.5 Flash \cite{nanobanana}) in Figure~\ref{fig:qual_comparison_appendix}. Our method consistently produces higher fidelity identity preservation and fewer geometric artifacts than competing approaches, particularly for challenging poses and diverse identities. Our method performs equally well on both in-domain (DeepFashion \cite{deepfashion}) and in-the-wild (WPose \cite{Li_UniHuman_2024}) scenarios, maintaining consistent quality across datasets.

\vspace{-2mm}

\subsection{Comparison with Gemini 3 Pro Image and Lighting Analysis}
\label{ssec:gemini3}
We specifically highlight a few examples in Figure~\ref{fig:gemini-3} which shows representative qualitative results of our method alongside Gemini 3 Pro Image~\cite{nanobananapro} and Gemini 2.5 Flash Image~\cite{nanobanana}. While the Gemini family produces visually compelling images, both models struggle with faithful reposing to the target SMPL-X pose, often failing to match the requested head and body pose. In contrast, Pro-Pose follows the target pose accurately while preserving identity. The same figure also illustrates our lighting behavior: because our Base Clothing standardization either re-lights subjects into even studio shading or matches the source illumination, Pro-Pose renders realistic, consistent lighting under novel poses without explicitly modeling target-pose illumination.

\begin{figure}[h]
    \centering
    \includegraphics[width=0.98\linewidth]{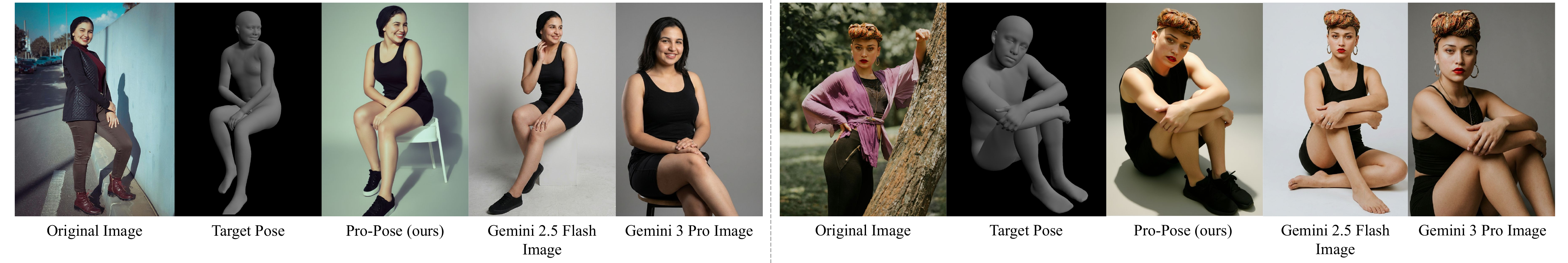}
    \vspace{-1mm}
    \caption{\textbf{Comparison with Gemini 3 Pro Image~\cite{nanobananapro} and lighting analysis.} Columns show the original image, the target pose, Pro-Pose (ours), Gemini 2.5 Flash Image~\cite{nanobanana}, and Gemini 3 Pro Image. Pro-Pose follows the target pose more faithfully while preserving identity and rendering realistic, consistent illumination.}
    \vspace{-6mm}
    \label{fig:gemini-3}
\end{figure}

\subsection{Evaluation with BC Inputs for Baselines}
\label{ssec:bc_baseline_eval}

A key difference between Pro-Pose and prior methods is that our model is trained on Base Clothing (BC) standardized inputs, while all baselines are trained on original images with diverse garments. To ensure a fair comparison, we additionally evaluate all baselines when provided with BC-preprocessed input images at test time. This removes garment variation as a confounding factor and isolates each method's ability to preserve identity and follow the target pose.

\noindent \textbf{Quantitative Results.}
Table~\ref{tab:merged-ablation-quant-suppl} reports the results. Even when baselines receive the same BC inputs that align with our training distribution, Pro-Pose maintains a significant advantage across identity (FaceSim) and perceptual quality (HPSv3) metrics, confirming that our gains stem from the method itself rather than the data preprocessing.

\noindent \textbf{Qualitative Results.}
Figure~\ref{fig:bc_baseline_qual} shows representative examples. Despite receiving the same standardized inputs, baselines still exhibit characteristic failures: identity drift (CFLD, MCLD), geometric artifacts in extreme poses (OnePoseTrans), and over-smoothed facial details (UniHuman). Pro-Pose consistently produces sharper, more identity-faithful results.

\vspace{-2mm}
\subsection{Extended Quantitative Evaluation}
\label{ssec:extended_quant}

To further validate the fidelity of our generated avatars, we report metrics computed exclusively on the facial regions in Table~\ref{tab:face-only-quant-table}. This complements the full-body metrics in the main paper by providing a stricter assessment of identity and facial detail preservation, where the large, uniform areas of background and body cannot inflate pixel-level scores.

\vspace{1mm}
\noindent \textbf{Metric Selection.}

\begin{wrapfigure}{r}{0.59\textwidth}
    \centering
    \vspace{-\intextsep}
    \resizebox{\linewidth}{!}{%
    \small
    \begin{tabular}{l|cccc}
    \hline
    \multicolumn{5}{c}{\textbf{DeepFashion (In-Domain)}} \\ \hline
    Method & \textbf{PSNR}$\uparrow$ & \textbf{SSIM}$\uparrow$ & \textbf{LPIPS}$\downarrow$ & \textbf{DINO}$\uparrow$ \\ \hline
    CFLD \cite{CFLD} & 14.34 & 0.7012 & 0.1823 & 0.9543 \\
    MCLD \cite{MCLD} & 15.76 & 0.7151 & 0.1798 & 0.9429 \\
    LEFFA \cite{Zhou2025LEFFA} & 13.95 & 0.6956 & 0.12 & 0.9316 \\
    OnePoseTrans \cite{oneposetrans} & 13.12 & 0.5738 & 0.322 & 0.9319 \\
    UniHuman \cite{Li_UniHuman_2024} & 13.87 & 0.7557 & 0.161 & 0.9534 \\
    Gemini 2.5 Flash \cite{nanobanana} & 15.59 & 0.7107 & 0.1018 & 0.9531 \\ \hline
    Unpaired Only & 12.87 & 0.6766 & 0.2489 & 0.9001 \\
    Paired Only & \best{16.45} & \second{0.7765} & \best{0.087} & \best{0.9555} \\
    \textbf{Ours (Unpaired + Paired)} & \second{16.38} & \best{0.7785} & \second{0.089} & \second{0.9554} \\ \hline
    \multicolumn{5}{c}{\textbf{WPose (Out-of-Domain)}} \\ \hline
    Method & \textbf{PSNR}$\uparrow$ & \textbf{SSIM}$\uparrow$ & \textbf{LPIPS}$\downarrow$ & \textbf{DINO}$\uparrow$ \\ \hline
    CFLD \cite{CFLD} & 15.12 & 0.741 & 0.279 & 0.6109 \\
    MCLD \cite{MCLD} & 15.29 & 0.757 & 0.24 & 0.6108 \\
    LEFFA \cite{Zhou2025LEFFA} & 16.47 & 0.77 & 0.236 & 0.5541 \\
    OnePoseTrans \cite{oneposetrans} & 17.1 & \second{0.807} & 0.174 & 0.6719 \\
    UniHuman \cite{Li_UniHuman_2024} & 17.49 & 0.802 & 0.179 & \second{0.6814} \\
    Gemini 2.5 Flash \cite{nanobanana} & 16.96 & 0.781 & 0.198 & 0.6776 \\ \hline
    Unpaired Only & 15.81 & 0.748 & 0.255 & 0.6445 \\
    Paired Only & \second{17.98} & 0.7866 & \second{0.165} & 0.6443 \\
    \textbf{Ours (Unpaired + Paired)} & \best{19.57} & \best{0.832} & \best{0.138} & \best{0.7043} \\ \hline
    \end{tabular}%
    }
    \vspace{-2mm}
    \captionof{table}{\textbf{Extended Quantitative Evaluation.} Complementing the full-body metrics in the main paper, we compute PSNR, SSIM, LPIPS, and DINO for the face region.}
    \label{tab:face-only-quant-table}
    \vspace{-\intextsep}
\end{wrapfigure}

We focus on PSNR, SSIM, LPIPS, and DINO to assess reconstruction quality and semantic alignment. We exclude OKS from this specific analysis, as it relies on full-body keypoints and becomes unstable when restricted to the sparse keypoints available in tight facial crops. Similarly, we omit HPSv3, as it is designed to assess global image aesthetics; when applied to small facial crops, it lacks the necessary global context to yield meaningful discriminative trends. Note that FaceSim is excluded from this table as it was already reported in the main paper (where it is inherently computed on facial embeddings).

\noindent \textbf{Analysis.}
The results reinforce the generalization gap observed in our main experiments. On the in-domain DeepFashion dataset, the "Paired Only" model performs marginally better, consistent with its tendency to overfit to the training distribution. However, on the challenging, out-of-distribution WPose dataset, Pro-Pose significantly outperforms all baselines and the "Paired Only" variant across all metrics. This confirms that our hybrid training strategy yields superior generalization on fine-grained identity features, avoiding the identity degradation observed in baselines.

\begin{figure}[htbp!]
  \centering
  \includegraphics[width=\linewidth]{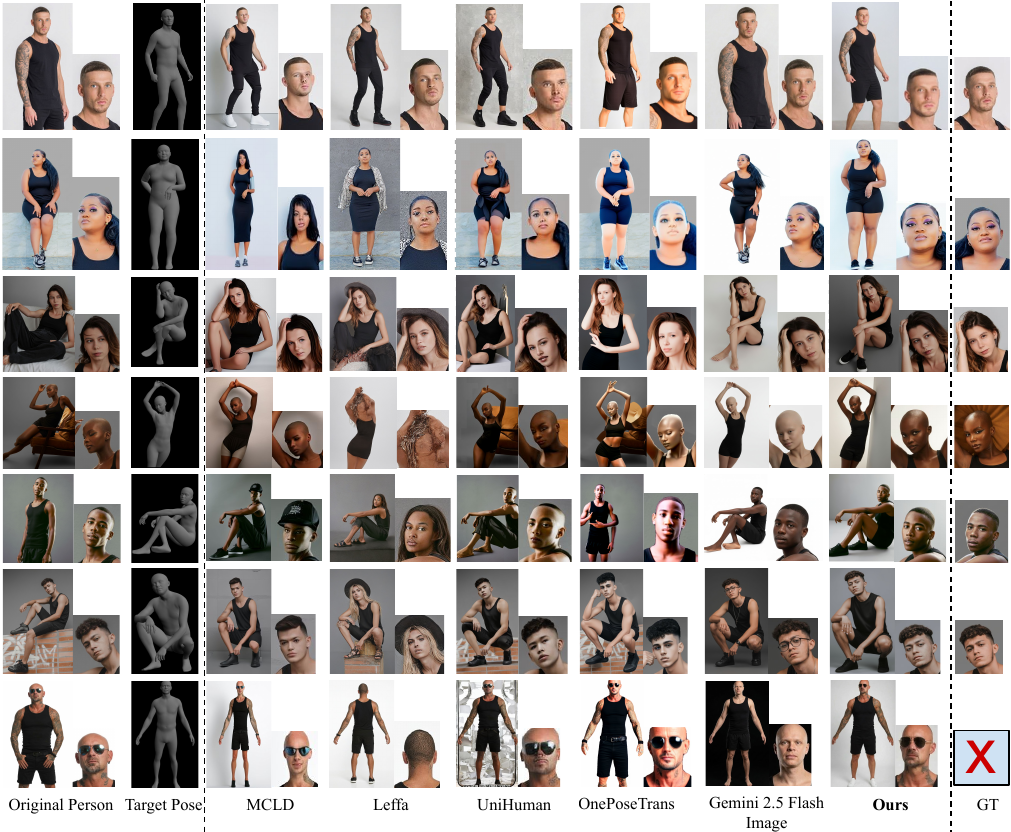}
  \caption{\textbf{Qualitative Comparison with BC Inputs.} All methods receive BC-preprocessed images as input, isolating the effect of the generation method from the data preprocessing. Pro-Pose preserves identity and pose fidelity more consistently than all baselines. For the final row, no ground truth (GT) exists because the input is a single image reposed into a target A-pose. Although a GT face crop cannot be provided, comparing the result to the face crop of the original image demonstrates our strong identity preservation.}
  \label{fig:bc_baseline_qual}
\end{figure}

\subsection{Full-Body Identity: Skin and Body-Shape Preservation}
\label{ssec:fullbody_identity}
While the main paper evaluates identity primarily through facial metrics, we additionally quantify how well Pro-Pose preserves \textit{body-skin appearance} and \textit{body shape}, directly assessing the full-body identity claim. Table~\ref{tab:shape-skin-metrics}A reports image and perceptual metrics computed exclusively on segmented body-skin pixels (excluding face, hair, and clothing). Table~\ref{tab:shape-skin-metrics}B measures body-shape consistency via the SMPL-X shape parameter $\hat{\beta}$, comparing the target and generated bodies using L2 distance and cosine similarity. As shown, Pro-Pose outperforms both OnePoseTrans~\cite{oneposetrans} and Gemini 3 Pro Image across all metrics, confirming that our gains extend beyond the face to whole-body skin and shape fidelity.

\begin{table}[h]
\centering
\vspace{-2mm}
\setlength{\tabcolsep}{4pt}
\resizebox{\linewidth}{!}{%
\begin{tabular}{l|cccc|cc}
\toprule
\textbf{WPose data} & \multicolumn{4}{c|}{\textbf{A. Body-skin preservation}} & \multicolumn{2}{c}{\textbf{B. Body-shape preservation}} \\
Method & PSNR$\uparrow$ & SSIM$\uparrow$ & LPIPS$\downarrow$ & DINO$\uparrow$ & $\hat{\beta}$ L2 $\downarrow$ & $\hat{\beta}$ Cosine-Similarity $\uparrow$ \\ \midrule
OnePoseTrans~\cite{oneposetrans} & 18.01 & 0.821 & 0.105 & 0.7292 & 2.156 & 0.6584 \\
Gemini 3 Pro Image & 19.13 & 0.841 & 0.091 & 0.7119 & 1.981 & 0.6921 \\
\textbf{Ours} & \textbf{22.34} & \textbf{0.892} & \textbf{0.079} & \textbf{0.7456} & \textbf{0.892} & \textbf{0.9487} \\
\bottomrule
\end{tabular}%
}
\vspace{1mm}
\caption{\textbf{Full-body identity preservation on WPose.} \textbf{A.} Metrics on body-skin pixels only (excluding face, hair, and clothing). \textbf{B.} SMPL-X $\hat{\beta}$ L2 distance and cosine similarity between target and generated bodies.}
\vspace{-10mm}
\label{tab:shape-skin-metrics}
\end{table}

\subsection{Impact of Training Data Source}
Complementing the quantitative ablation in the main paper, Figure~\ref{fig:ablation_training_data_sources_appendix} visually demonstrates the impact of our hybrid data strategy. Models trained on "Unpaired Only" data struggle with extreme poses, while "Paired Only" models suffer from identity drift. Our combined approach ("Unpaired + Paired") leverages the strengths of both, achieving robust pose control and identity preservation.

\begin{figure}[htbp!]
  \centering
  \includegraphics[width=0.98\linewidth]{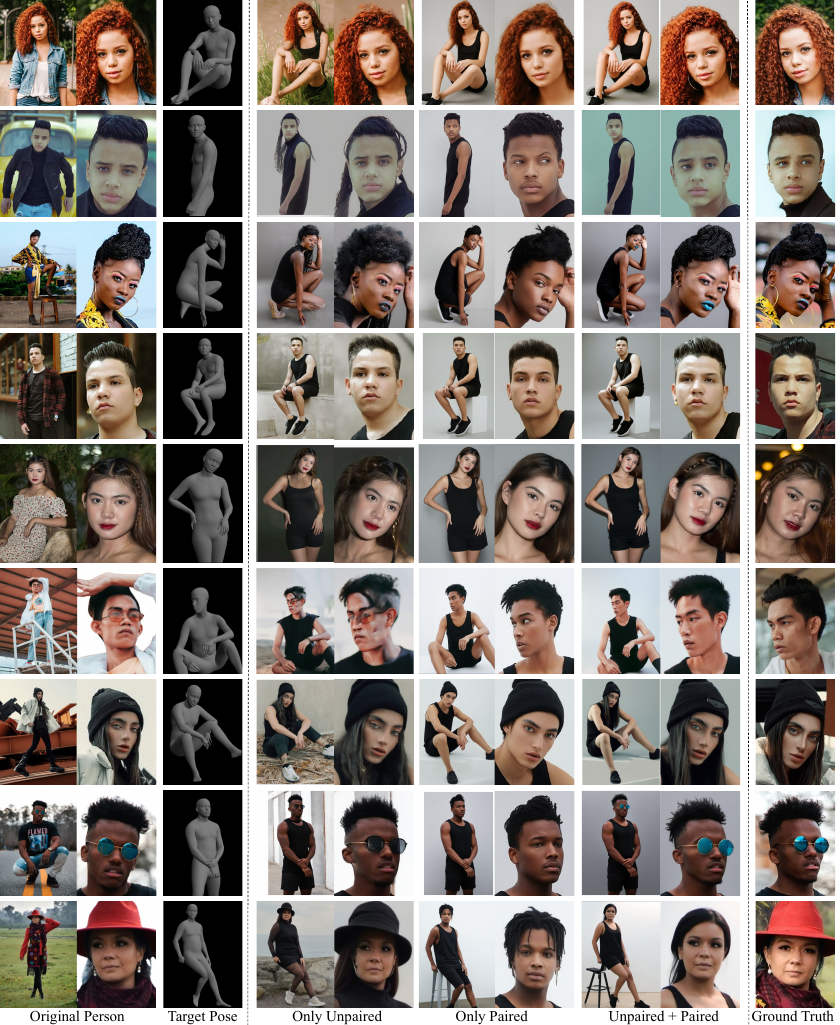}
  \caption{\textbf{Ablation Study of Training Data Sources.}
  We evaluate the impact of different training dataset combinations on generated avatar quality. The result columns display models trained on Unpaired data only, Paired data only, and the full Unpaired + Paired combination, respectively. One can observe the improvement in fidelity with the full combined dataset.
  }
  \label{fig:ablation_training_data_sources_appendix}
\end{figure}

\subsection{Ablation: UV as the Control Signal}
\label{ssec:uv_ablation}
A natural question is whether Pro-Pose's improvements stem specifically from using the canonical UV texture as the control signal. In the unpaired regime this cannot be ablated, as the UV map is the \textit{sole} source of texture information; omitting it leaves the self-supervised objective undefined. We therefore ablate the UV condition in the \textit{paired-only} setting, where a face crop still provides identity information. As shown in Table~\ref{tab:UV_ablation} (on WPose, directly comparable to Table~1 of the main paper), removing the UV map from paired-only training yields only a modest degradation and closely matches the paired-only baseline with UV. This confirms that our primary gains stem not from the UV control signal alone, but from our novel use of large-scale unpaired data via donor-based reposing.

\begin{table}[h]
\centering
\resizebox{\linewidth}{!}{%
\begin{tabular}{l|cccccccc}
\toprule
Pro-Pose (paired only) & \textbf{M-PSNR}$\uparrow$ & \textbf{FID}$\downarrow$ & \textbf{M-SSIM}$\uparrow$ & \textbf{M-LPIPS}$\downarrow$ & \textbf{OKS}$\uparrow$ & \textbf{FaceSim}$\uparrow$ & \textbf{DINO}$\uparrow$ & \textbf{HPSv3}$\uparrow$ \\ \midrule
with UV & 18.30 & 6.65 & 0.820 & 0.155 & 0.34 & 0.4959 & 0.6972 & 7.30 \\
without UV & 18.05 & 7.11 & 0.801 & 0.192 & 0.34 & 0.4721 & 0.6888 & 6.98 \\
\bottomrule
\end{tabular}%
}
\vspace{1mm}
\caption{\textbf{Ablation of UV as the control signal.} Removing the UV map from paired-only training (comparable to Table~1, main paper, WPose) produces only a modest drop, confirming that Pro-Pose's gains stem primarily from our use of unpaired data rather than the UV control signal alone.}
\vspace{-7mm}
\label{tab:UV_ablation}
\end{table}

\subsection{Downstream Virtual Try-On: Same-Pose Comparison}
\label{ssec:vto_suppl}
To evaluate Pro-Pose's benefit as a VTO pre-processing stage under a fair, pose-matched setting, Figure~\ref{fig:vto_example} compares applying an off-the-shelf VTO model~\cite{googlevto2023} directly to the original in-the-wild image versus applying it to the Pro-Posed avatar, with both branches targeting the same pose. VTO on the original image frequently fails due to source-garment interference, whereas VTO on the Pro-Posed avatar succeeds. This qualitatively complements the user study reported in the main paper (Section 4.4).

\begin{figure}[h]
    \centering
    \vspace{-4mm}
    \includegraphics[width=\linewidth]{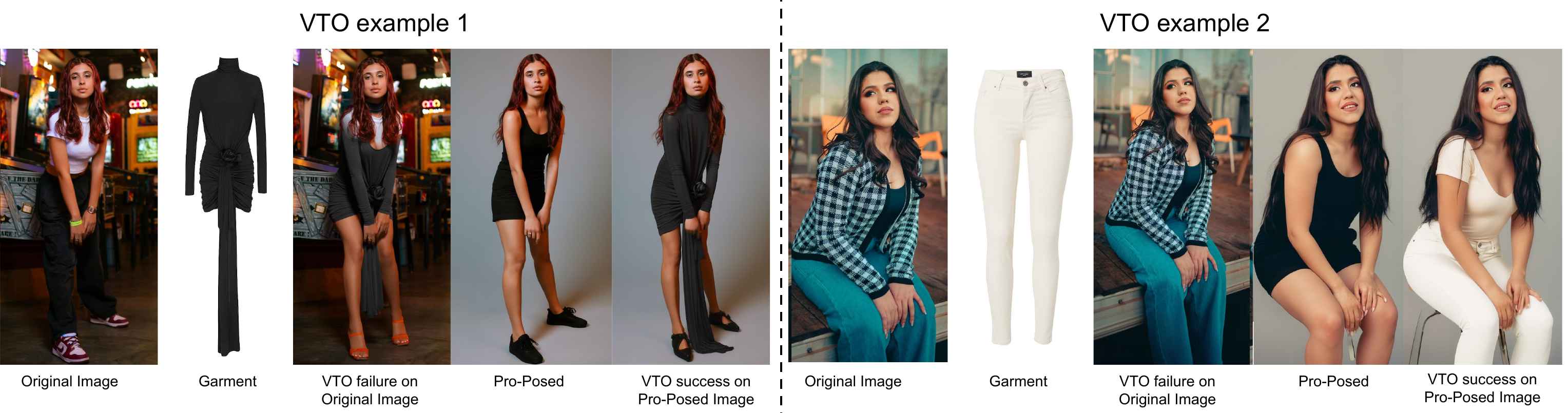}
    \caption{\textbf{Virtual Try-On under matched target pose.} Despite the same target pose, VTO fails on original images due to source-garment interference, but succeeds on Pro-Posed avatars.}
    \vspace{-8mm}
    \label{fig:vto_example}
\end{figure}

\subsection{Personalization with Few-Shot Learning}
\label{ssec:finetuning_appendix}

As outlined in Section 3.5 of the main paper, we perform few-shot adaptation to specialize the model for a specific target identity. Here, we provide the specific implementation details for this process.

\noindent \textbf{Data Source and Preprocessing.} 
While the visualization (Figure 4) in the main paper utilizes frames from a video sequence, our fine-tuning protocol is data-agnostic. It requires only a set of paired images of the subject in different poses; these can be sourced from a video or a collection of independent multi-view photographs.
Crucially, before fine-tuning, we process all reference images to match our \textbf{Base Clothing (BC)} standard (black tank top and shorts). This ensures that the input data aligns with the canonical distribution the model was trained on, allowing the optimization to focus on identity features rather than clothing discrepancies.

\noindent \textbf{Implementation Details.} 
We fine-tune the pre-trained LoRA adapters for 5,000 iterations with a batch size of 4. To prevent the model from overfitting to the limited few-shot examples or forgetting its general geometric priors, we include a small set of regularization images in the training batches.
For consistency and reproducibility, we maintain all other hyperparameters (including learning rate and optimizer settings) identical to the main training phase. While subject-specific hyperparameter tuning could theoretically yield higher fidelity results, we report all metrics using this fixed configuration.

\begin{figure}[t]
    \centering
    \includegraphics[width=1\linewidth]{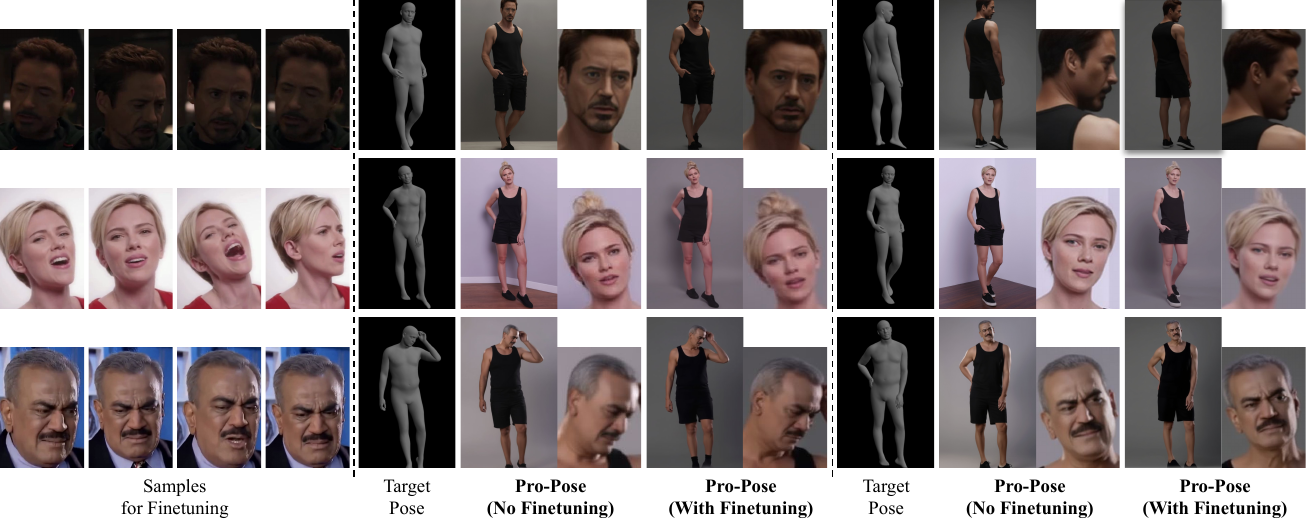}
    \caption{\textbf{Personalized Pro-Pose.} We fine-tune our method using the images shown in the leftmost column (rows 1-4). Column (5) displays the target pose represented by a SMPL-X mesh (reposed using ground truth parameters). Columns (6) and (7) present the generated output without and with fine-tuning, respectively. The subsequent columns repeat this arrangement for a second target pose.}
    \label{fig:finetuning_appendix_qual}
\end{figure}

\noindent \textbf{More Qualitative Evaluation.}
We demonstrate how our few-shot learning strategy generalizes to in-the-wild scenarios~\cite{celebVHQ} in Figure~\ref{fig:finetuning_appendix_qual}. Notably, the identities of the generated new poses become significantly closer to the original subject after fine-tuning, validating the effectiveness of this adaptation step.

\begin{figure}[t]
    \centering
    \includegraphics[width=0.98\linewidth]{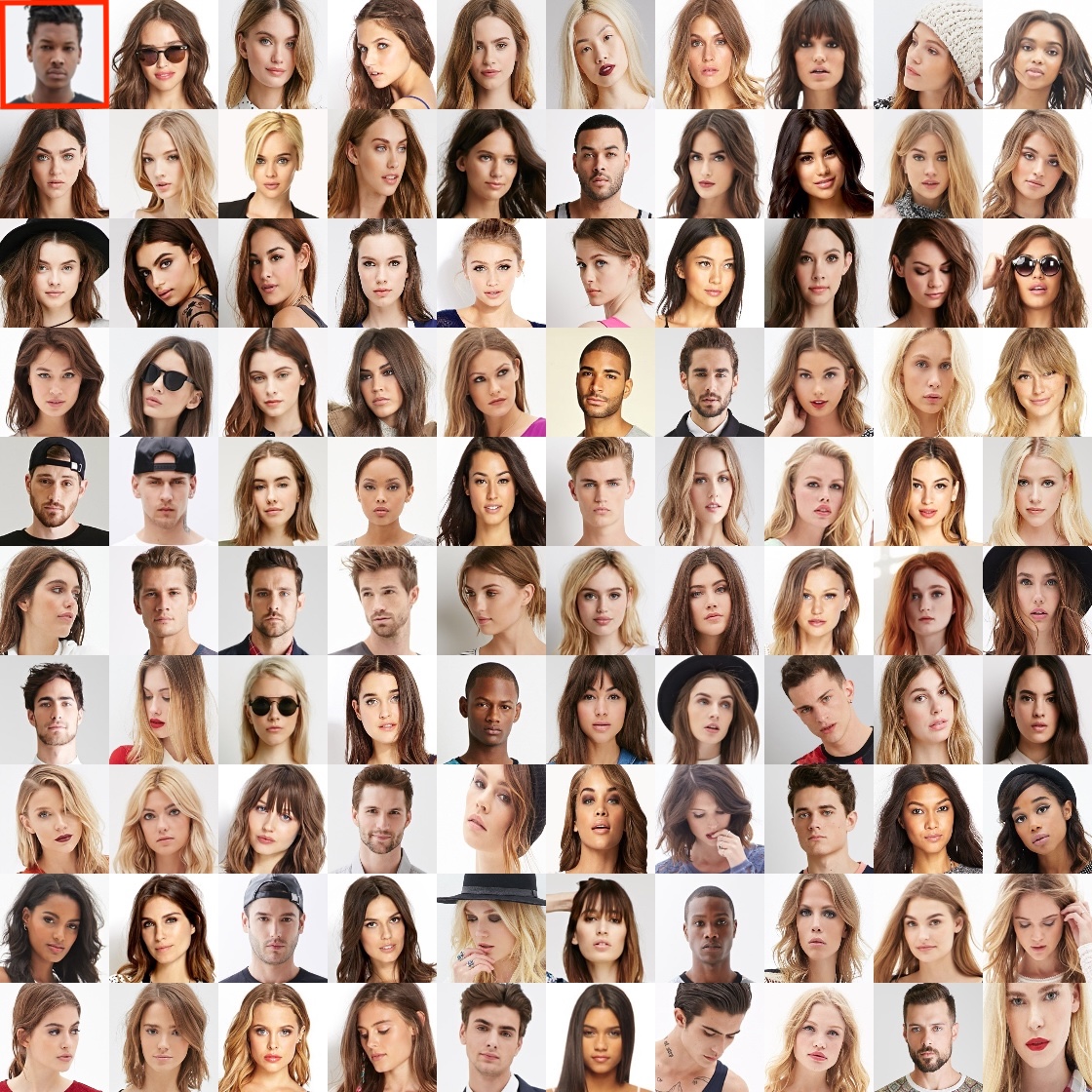}
    \caption{\textbf{DeepFashion.} We highlight the top 100 identity clusters found in the DeepFashion dataset. We also exclusively highlight the first person on the top-left, who can be repeatedly observed in the generations of the overfitted models.}
    \label{fig:deepfashion_100_identities}
\end{figure}

\begin{figure}[t]
    \centering
    \includegraphics[width=0.98\linewidth]{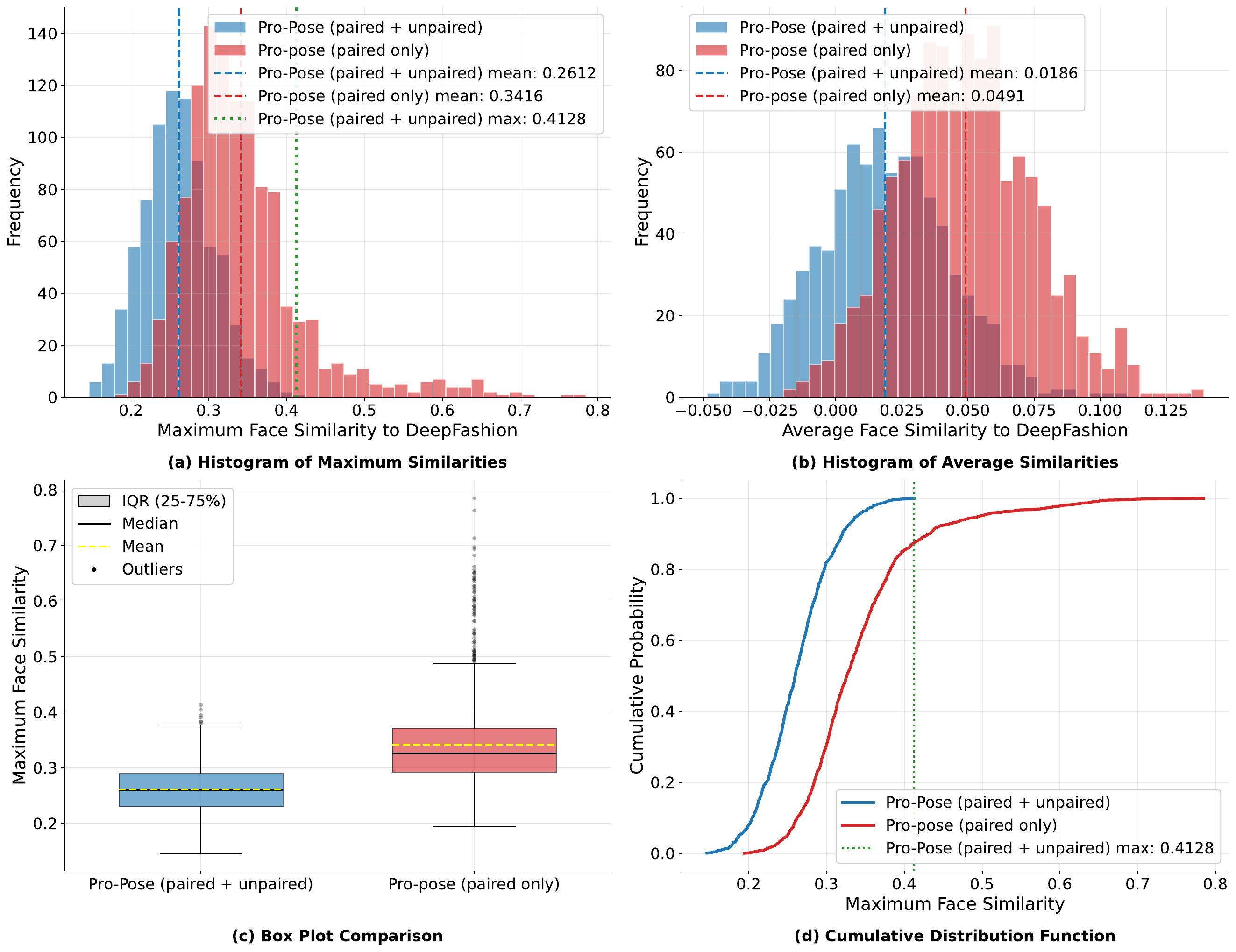}
    \caption{\textbf{Diagnosing Identity Overfitting via Cross-Dataset Generation.} We evaluate the extent of identity overfitting by comparing the face similarity between images generated on the out-of-distribution WPose dataset and the identities present in the DeepFashion training set. Two Pro-Pose models are tested: Pro-Pose (paired only), trained on limited DeepFashion paired data, and Pro-Pose (unpaired + paired), trained on abundant unpaired data plus the limited paired data. The Pro-Pose (paired only) model exhibits a significantly higher face similarity to the DeepFashion training identities when reposing WPose identities. This result acts as a diagnostic, demonstrating that the paired-only model has overfit to the limited DeepFashion identities, causing it to generate new faces that are biased towards its training set, whereas the combined training approach ensures better generalization.}
    \label{fig:paired_Data_bias_analysis}
\end{figure}

\section{Analysis of Identity Overfitting}
\label{sec:overfitting_analysis}
A primary motivation for our work is the limited identity diversity in existing paired datasets like DeepFashion. As visualized in Figure~\ref{fig:deepfashion_100_identities}, DeepFashion contains limited ($\approx 100$) unique identities. Training generative models exclusively on such limited data leads to severe overfitting, where the model biases generated faces toward the training identities rather than the input subject.

\subsection{DeepFashion Identity Analysis}
\label{sec:identity_analysis}

To empirically validate the limited identity diversity of the DeepFashion In-Shop Clothes Retrieval benchmark~\cite{deepfashion}, we performed a graph-based clustering analysis on the training set. Our pipeline proceeds as follows:

\begin{enumerate}
    \item \textbf{Embedding Extraction:} We utilized ArcFace~\cite{arcface} to detect faces and extract identity embeddings. To ensure embedding reliability, we explicitly filtered out faces with extreme head poses (yaw or pitch $> 45^\circ$).
    \item \textbf{Graph Construction:} We constructed an undirected graph where nodes represent images. Edges were established via a dual-criteria strategy: 
    \begin{itemize}
        \item \textit{Visual Similarity:} An edge is created if the cosine similarity between two embeddings exceeds a strict threshold of $0.6$.
        \item \textit{Structural Verification:} We leveraged the dataset's directory structure (where images in the same garment folder typically depict the same subject) to add ground-truth edges. However, to filter labeling errors or occlusions, these structural edges were only added if the similarity exceeded a verification threshold of $0.4$.
    \end{itemize}
    \item \textbf{Clustering:} We computed the connected components of this graph. This process yielded approximately 100 distinct clusters (identities), confirming that the dataset is heavily dominated by a small number of professional models appearing across thousands of garment items.
\end{enumerate}

\subsection{Identity Analysis of Generated Images}
We diagnose this phenomenon in Figure~\ref{fig:paired_Data_bias_analysis}. We reposed subjects from the out-of-distribution - WPose dataset using two models: one trained only on paired DeepFashion data, and our full model trained on both paired and unpaired data. We then measured the face similarity between the \textit{generated} outputs and the \textit{DeepFashion training set}.
\begin{itemize}
    \item The \textbf{Paired-Only} model (red histograms in Figure~\ref{fig:paired_Data_bias_analysis} (a) and (b)) produces outputs with high similarity to the training data, indicating it is "pulling" novel identities toward the few identities it memorized during training.
    \item \textbf{Our Full Model} (blue histograms in Figure~\ref{fig:paired_Data_bias_analysis} (a) and (b)) maintains lower similarity to the training set, proving that the inclusion of abundant unpaired data enables better generalization to unseen identities.
\end{itemize}

The box-plot in Figure~\ref{fig:paired_Data_bias_analysis} (c) further highlights the extensive number of samples generated by the paired-only model that have very high similarity scores to DeepFashion training samples.

The cumulative distribution (Figure~\ref{fig:paired_Data_bias_analysis} (d)) highlights that our model trained on both paired and unpaired data generated samples with a maximum Face Similarity score of $0.4128$, while almost $15\%$ of the total samples generated by the paired only model scored an even higher face similarity score. 

\subsection{DeepFashion Identity Limitation Leading to Overfitting}
\label{ssec:identity_visual_proof}

We highlight a dominant identity from the DeepFashion training set in the red box of Figure~\ref{fig:deepfashion_100_identities}. Comparing this to the "Paired Only" results in Figure~\ref{fig:ablation_training_data_sources_appendix} (and Figure 7 of main paper) reveals severe overfitting: As we can see in some cases the model tends to collapse to this specific training identity (for example second row in Figure~\ref{fig:ablation_training_data_sources_appendix}). This bias is so strong that it overrides semantic gender cues, frequently substituting female subjects with this specific male face, confirming that the paired-only model relies on memorization rather than generalization.

\end{document}